\pdfoutput=1
\documentclass{article}

\PassOptionsToPackage{numbers, sort&compress}{natbib}
\usepackage[main, final]{neurips_2026}

\usepackage[utf8]{inputenc} 
\usepackage[T1]{fontenc}    
\usepackage{hyperref}        
\usepackage{url}            
\usepackage{booktabs}       
\usepackage{amsfonts}       
\usepackage{nicefrac}       
\usepackage{microtype}      
\usepackage{xcolor}         
\definecolor{NIPSblue}{RGB}{0,80,160}
\hypersetup{colorlinks=true, linkcolor=NIPSblue, citecolor=NIPSblue, urlcolor=NIPSblue, filecolor=NIPSblue}
\usepackage{graphicx}
\usepackage[normalem]{ulem}
\usepackage{amsmath}

\title{The Shape of Events: Edge-Based Inductive Biases via Cross-Domain Distillation}

\author{%
Soshun Kihara$^{1}$\thanks{All authors contributed equally.}
\quad
Shunsuke Yasuki$^{2}$\footnotemark[1]
\quad
Masato Taki$^{3,4}$\footnotemark[1] \\[3pt]
$^{1}$Independent Researcher \qquad $^{2}$The University of Tokyo, Japan \\
\makebox[\dimexpr\textwidth-2\tabcolsep\relax][c]{$^{3}$Graduate School of Artificial Intelligence and Science, Rikkyo University, Japan \quad $^{4}$RIKEN, Japan} \\[5pt]
{\small \texttt{project page:} \url{https://github.com/snskysk/event2rgb-distillation}}
}

\begin{document}

\maketitle
\vspace{-8pt}


\begin{abstract}

Convolutional neural networks trained on ImageNet are known to exhibit a strong preference for local high-frequency texture, an inductive bias that translates into fragile robustness against distribution shifts in real-world environments. Event cameras, in contrast, record only changes in scene brightness and are therefore well suited to capturing contour information; however, due to the absence of diagnostic benchmarks in the event domain, the inductive bias that event-camera data instills in vision models has remained underexplored. In this work, we use knowledge distillation from the event domain to the RGB domain so as to exploit the rich evaluation toolkit available in the RGB domain and systematically dissect this inductive bias. Our experiments show that distillation from the event domain induces, in the RGB domain, color invariance, shape bias, and robustness to high-frequency noise. We identify the underlying mechanism as the model suppressing its dependence on high-frequency texture while acquiring a stronger dependence on edge-based object shape. This hypothesis is supported by changes in how color and spatial information are processed at the early layers, together with a spectral trade-off in which robustness to the absence of high-frequency components coexists with vulnerability to contamination of the relied-upon frequency bands and to disruption of geometric structure. We further show that this inductive bias differs from existing robustification methods and that it functions as a useful prior for diverse downstream tasks in which shape and contour information contribute alongside other cues.

\if[]
Standard computer vision models exhibit a strong reliance on local high-frequency texture, which is a fundamental cause of their vulnerability in real-world settings. Event-based cameras, by virtue of their physical characteristic of capturing object brightness changes (edges), should impart a distinctive inductive bias to models, yet the lack of diagnostic benchmarks in the event domain has left the precise nature of this bias unexplained. In this work, we propose a new framework that uses knowledge distillation from the event domain to the natural image (RGB) domain, leveraging the rich evaluation toolkit of the RGB domain to systematically dissect this inductive bias. Extensive validation reveals that Event-distilled models exhibit unusually strong color invariance, shape bias, and robustness to high-frequency noise.
We identify the mechanism as the model suppressing unstructured high-frequency texture and selectively extracting macroscopic ``edge-based shape feature,'' i.e.\ acquiring a ``Structural Prior.''
This hypothesis is corroborated by internal-representation changes including increased spatial correlation in the early layers, alongside the ``spectral trade-off'' woven from robustness to the loss of high-frequency components and vulnerability to contamination of the relied-upon bands and to structural disruption.
We further show that this inductive bias differs markedly from those induced by existing robustification methods, and
functions as a strong structural prior for diverse downstream tasks demanding shape-based reasoning, such as medical imaging.

\fi

\end{abstract}


\section{Introduction}
\label{sec:intro}

Unlike the human visual system, convolutional neural networks (CNNs) are widely reported to exhibit a texture bias, preferring local high-frequency texture over global object shape~\cite{geirhos2019texture_bias}. This inductive bias is an underlying cause of model fragility under real-world environmental changes such as illumination variation, noise, and domain shift. In this paper, we present a new approach to this problem from the perspective of the event-based sensor (event camera) domain.

Event cameras possess properties that conventional cameras lack, including high temporal resolution and high dynamic range~\cite{gallego2020event_survey}. Moreover, their hardware design records only per-pixel changes in brightness: uniform static regions emit no events, whereas spatial brightness gradients (including textured regions) generate events only when motion converts them into temporal changes. Edges and contours are therefore represented in the event stream more reliably than textures, whose presence is motion-dependent, and we expect models trained on event data to acquire internal representations grounded in macroscopic geometric structure and edges.

However, research on the internal representations of models trained on event data is still insufficient. In the RGB domain, benchmarks for dissecting and diagnosing models are well established, including evaluations against diverse corruptions such as ImageNet-C~\cite{hendrycks2019imagenetc} and shape--texture cue-conflict images~\cite{geirhos2019texture_bias}. In the event-vision community, by contrast, evaluation metrics and diagnostic datasets that allow detailed analysis of model behavior are lacking. As a result, the representations that event data induces in vision models remain underexplored.

To overcome this evaluation barrier and dissect the representations specific to event data, we propose an approach based on the simple framework of knowledge distillation~\cite{hinton2015knowledge_distillation}. Specifically, we use a model trained on the large-scale event dataset Neuromorphic-ImageNet (N-ImageNet)~\cite{kim2021nimagenet} as a teacher, and transfer (distill) its knowledge into a student model in the RGB domain (ImageNet~\cite{russakovsky2015imagenet}). This cross-domain distillation has two implications. First, by transplanting the features and properties of event data into the RGB domain, we can dissect their underlying mechanism using the rich evaluation toolkit of the RGB domain. Second, by distilling event-derived constraints into an RGB model, where texture would otherwise dominate as a shortcut, we can suppress shortcut learning in a way that has not been studied before. Note that, because N-ImageNet is generated by recapturing ImageNet on a monitor, our experiments isolate the brightness-gradient (edge) selectivity property of event data; HDR and motion parallax are outside the experimental scope (Section~\ref{sec:limitations}).

Our experiments reveal that models distilled with knowledge from event data (hereafter, Event-distilled models) acquire a set of robustness properties (color invariance, shape bias, and resistance to high-frequency noise) compared to baseline models. Furthermore, by analyzing the high spatial correlation of input-layer filters and the model's behavior under digital artifacts (robustness to JPEG compression, in which unstructured texture is removed, and vulnerability to perturbations that destroy geometric structure such as edge continuity), we examine the mechanism behind this robustness. That is, the Event-distilled model learns an edge-based shape feature, a property distinct from prior robust models and shape-biased models.

The main contributions of this work are summarized as follows:

\begin{itemize}
    \item Event-driven knowledge distillation: We propose an analytical use of cross-domain distillation from event- to RGB-trained models, deliberately adopting the standard knowledge-distillation formulation~\cite{hinton2015knowledge_distillation}. Our contribution lies not in a new distillation objective but in the cross-domain setup and in the systematic characterization of the resulting inductive bias through RGB-domain diagnostics.
    \item A characteristic set of robustness properties: Through comprehensive evaluation, we observe that the Event-distilled model exhibits a set of phenomena, including color invariance, a shift toward geometric and structural cues over texture, and improved accuracy under high-frequency noise and under adversarial perturbations with $0 < \epsilon \le 1/255$.
    \item Mechanism elucidation and downstream applications: We identify that the source of these robustness properties is a shift in the internal representation, in which the model suppresses its dependence on high-frequency texture and acquires a stronger dependence on the edge-based shape feature. We further show that this property functions as a useful inductive bias for downstream tasks, suggesting new directions for event data in computer vision.
\end{itemize}

\begin{figure*}[!t]
\vspace{-1.5em}
\centering
\includegraphics[width=\linewidth]{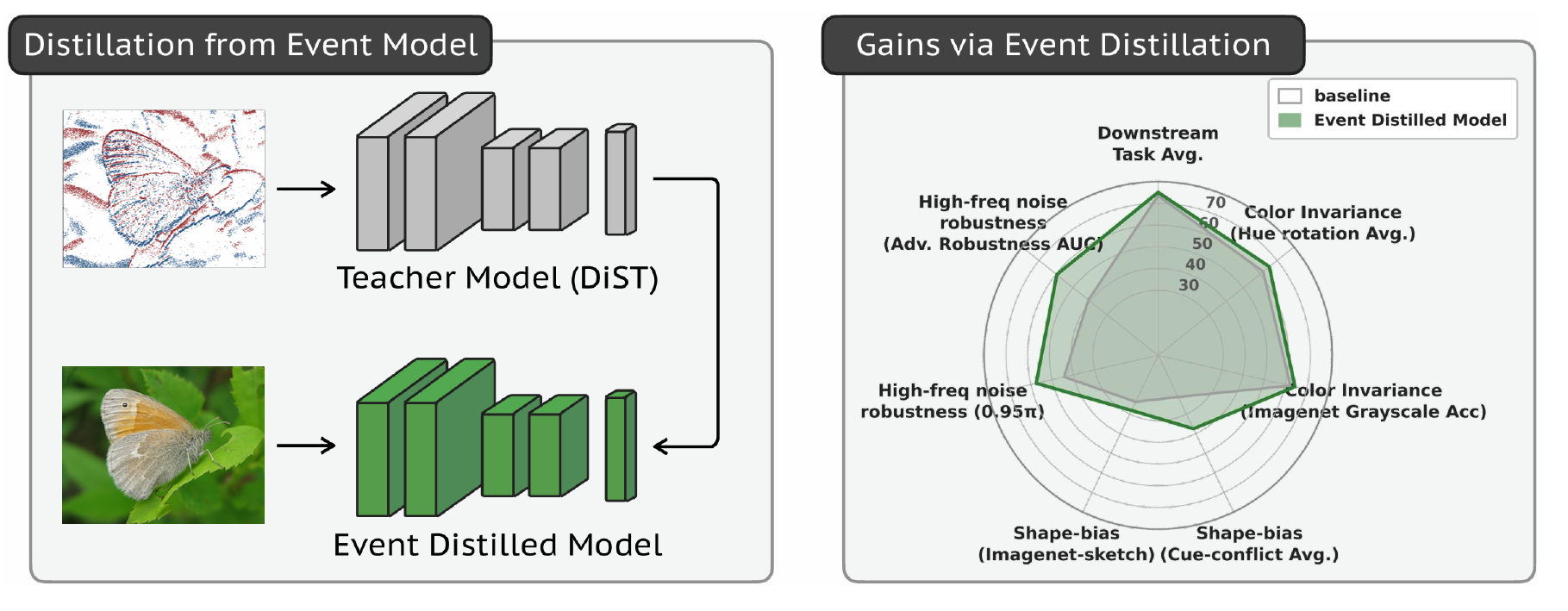}
\vspace{-1.1em}
\caption{Method overview and summary of results.
\emph{Left}: Cross-domain knowledge distillation from a frozen DiST event teacher (trained on N-ImageNet) into an RGB ResNet-34 student (trained on ImageNet). N-ImageNet and ImageNet are paired at the instance level, enabling per-sample KL between teacher and student logits.
\emph{Right}: Summary of robustness gains of ED\_Student ($\alpha = 0.2$) over the RGB baseline across the three axes studied in this paper (color invariance, shape bias, and high-frequency noise robustness) and downstream task transfer.}
\label{fig:method_overview}
\vspace{-1em}
\end{figure*}

\if[]

Text not to be used

\fi

\section{Related Work}
\label{sec:related_work}

\subsection{Event-based Vision and Neuromorphic Datasets}

Event cameras asynchronously capture brightness changes~\cite{gallego2020event_survey} and are robust to illumination change and motion blur.
The recent emergence of large-scale datasets such as N-ImageNet~\cite{kim2021nimagenet,orchard2015converting,serrano2015poker,hu2016dvs,li2017cifar10,amir2017low,lungu2017live,sironi2018hats,moeys2018pred18,bi2019graph,cheng2019det,vasudevan2020introduction,ramesh2020low} has driven the proposal of recognition models such as DiST~\cite{kim2021nimagenet}, including a wide range of subsequent approaches~\cite{kim2022ev,cho2023label,wu2023eventclip,klenk2024masked,cao2024spiking,yang2025ezsr,kowalczyk2025learning,fan2025eventpillars,liang2025efficient,zhou2024eventbind,xu2024ceia,jeong2026cross,yan2026omnievent,ma2026i2e}; following N-ImageNet, many of them convert the event stream into a grid representation (an event image) so that existing architectures such as CNNs can be reused.
Applications to other computer-vision tasks such as object detection~\cite{perot2020learning,mondal2021moving,gehrig2023recurrent,zubic2023chaos,gehrig2024low,zubic2024state,yang2025smamba} and semantic segmentation~\cite{alonso2019ev,sun2022ess,hamaguchi2023hierarchical,kong2024openess,li2025efficient} are also expanding.

To compensate for the scarcity of labeled event data, distillation from a pretrained RGB model into an event model has been proposed~\cite{wang2021evdistill,yang2023event,zhang2025adaptive,wu2026ewad}.
To our knowledge, no work sends knowledge in the opposite direction, from an event model into an RGB model, in order to analyze the representational properties that event data induces.
Hybrid sensors such as DAVIS~\cite{brandli2014davis} read out frames and events from the same pixel array; as natively paired data become available at scale, our dissection could be repeated on them.

\subsection{Inductive Biases in CNNs: Texture vs.\ Shape}

It has been widely pointed out that standard CNNs trained on RGB images exhibit a texture bias, preferring local high-frequency texture over the overall shape of objects~\cite{geirhos2019texture_bias,baker2018deep,geirhos2018generalisation,brendel2019bagnet,hermann2020origins,islam2021shape,mummadi2021does,li2021shapetexture,geirhos2021cue_conflict}. In response, shape-biased models have been trained on style-transferred images such as Stylized-ImageNet~\cite{geirhos2019texture_bias}, which destroy texture and force the model to learn shape.
Adversarial training and noise-augmentation training are also known to reduce texture reliance as a side effect. The Event-distilled model likewise exhibits a strong shape bias, but we show in Sections~\ref{sec:6-3_downstream} and \ref{sec:6-4_vs_robust_methods} that it is grounded in an edge-based shape feature distinct from those of Adversarial Training, Noise Training, and Stylized-ImageNet co-training (hereafter, the SIN model)~\cite{geirhos2019texture_bias}.

\subsection{Knowledge Distillation for Model Analysis}

Knowledge distillation is a widely used technique that trains a student model using the predictive distribution (soft labels) of a teacher model~\cite{hinton2015knowledge_distillation,romero2015fitnets,park2019rkd,tian2020crd,touvron2021deit,beyer2022patient}. More recently, cross-modality distillation that transfers representations between different modalities (e.g., RGB and depth, RGB and audio) has also been investigated~\cite{gupta2016cross,aytar2016soundnet,hoffman2016hallucination,garcia2018modality,zhang2025adaptive,wu2026ewad}.
Our goal, however, is neither compression nor accuracy on a new task. Because diagnostic metrics are underdeveloped in the event domain, we transfer knowledge from an N-ImageNet teacher into an RGB student that has access to abundant diagnostics, repurposing distillation as an analytical tool for dissecting the inductive bias of a heterogeneous domain.

\section{Method: Distilling Event-Driven Inductive Biases}
\label{sec:method}

Given the limited means available for directly analyzing the visual features (inductive biases) on which event-domain models rely, we propose a framework that repurposes cross-domain distillation as an analytical tool. Figure~\ref{fig:method_overview} provides an overview. Concretely, we adopt the DiST ResNet-34~\cite{kim2021nimagenet} trained on N-ImageNet as the teacher model, and a ResNet-34 trained on ImageNet~\cite{russakovsky2015imagenet} as the student model.

\textbf{Knowledge distillation (KD) preliminaries.} Knowledge distillation~\cite{hinton2015knowledge_distillation} trains a student network to match the temperature-softened softmax outputs of a teacher network. In our setup the teacher is trained on event-domain data (N-ImageNet via DiST~\cite{kim2021nimagenet}), while the student learns RGB classification with distillation guidance from the teacher's event-domain predictions on paired samples.

\subsection{Model Design and Training Strategy}

To analyze the inductive bias with a classical CNN structure while ensuring fair comparison with prior work (DiST~\cite{kim2021nimagenet}), we adopt ResNet-34~\cite{he2016resnet} as our principal subject of analysis. We verify the architectural generality of our findings in Section~\ref{sec:6-2_architectural_generality}, using VGG~\cite{simonyan2015vgg}, ConvNeXt~\cite{liu2022convnext}, and others. We choose ResNet-34 rather than the more common ResNet-50 for three reasons: the DiST teacher is itself a ResNet-34, which gives a same-architecture teacher--student pair; its moderate size (about 21M parameters) keeps the teacher variants and the multiple students within our compute budget; and the phenomena replicate on a ResNet-50 student (Section~\ref{sec:6-2_architectural_generality}), the backbone used in the original cue-conflict study~\cite{geirhos2019texture_bias}, so this choice does not restrict our conclusions.

As in standard knowledge distillation~\cite{hinton2015knowledge_distillation}, the student model is trained to minimize a linear combination of the cross-entropy loss ($\mathcal{L}_{CE}$) with the ground-truth labels (hard label) and the Kullback--Leibler divergence ($\mathcal{L}_{KL}$) with the teacher model's predictive distribution (soft label). The overall loss $\mathcal{L}$ in this work is defined using a weight parameter $\alpha \in [0, 1]$ as follows.

\begin{equation}
    \mathcal{L} = (1 - \alpha)\,\mathcal{L}_{CE}\bigl(y,\, p_s(x_{\text{strong}})\bigr) + \alpha\,T^2\,\mathcal{L}_{KL}\bigl(p_t(x_{\text{event}}),\, p_s(x_{\text{weak}})\bigr)
    \label{eq:eq001}
\end{equation}

Here, $y$ is the ground-truth label, $x_{\text{strong}}$ and $x_{\text{weak}}$ denote the same RGB instance passed through a strong and a weak augmentation pipeline respectively, $x_{\text{event}}$ is the paired event sample, and $p_s$ and $p_t$ are the (softmax) outputs of the student and teacher models. $\mathcal{L}_{KL}$ uses temperature-softened logits ($T = 3.0$) with the standard $T^2$ multiplier~\cite{hinton2015knowledge_distillation}. This dual-branch construction follows DeiT~\cite{touvron2021deit} and Patient Teacher~\cite{beyer2022patient}; see Appendix~\ref{sec:ap_kd_view_alignment} for the augmentation pipelines and cross-modal view alignment. To isolate the influence of the event-derived inductive bias on the RGB model, in addition to a baseline model (a ResNet-34 trained on ImageNet RGB images, i.e.\ $\alpha = 0$), we design the following two Event-distilled variants by varying the soft-label dependence $\alpha$.

\begin{itemize}
    \item ED\_SoftLabelOnly ($\alpha = 1.0$): A student model trained using only the teacher's soft labels (soft:hard = 10:0). By completely eliminating the intervention of label information from the RGB domain, this model is expected to inherit the predictive distribution (inductive bias) of the event teacher in its purest form. However, because the DiST ResNet-34 teacher itself attains only 48.43\% on N-ImageNet, the standard accuracy on clean ImageNet drops considerably from the baseline (73.9\%$\rightarrow$53.5\%).
    \item ED\_Student ($\alpha = 0.2$): A model trained with a mixture of soft and hard labels (soft:hard = 2:8). To prevent large clean-accuracy drops from confounding our analysis, we adopt the empirically standard $\alpha = 0.2$ weighting that balances regularization and knowledge transfer, preserving the event-derived inductive bias while alleviating degradation of basic image-classification capability.
\end{itemize}

The notation soft:hard refers only to the $\alpha$ weighting in Eq.~\ref{eq:eq001} ($\alpha$ for the KL term and $1-\alpha$ for the CE term), before the standard $T^2$ scaling of the KL loss. With $T = 3$ the effective coefficient of the KL term is therefore $\alpha\,T^2 = 1.8$ for ED\_Student, but we keep the 2:8 notation for the $\alpha$ weighting itself, following the convention of prior knowledge-distillation work.

Since our objective is mechanism analysis rather than SOTA accuracy, we do not exhaustively tune the soft/hard label mixing ratio. Detailed training settings including data augmentation are described in Appendix~\ref{sec:ap_exp_setup}.

\begin{figure*}[!t]
\vspace{-1.5em}
\centering
\includegraphics[width=\linewidth]{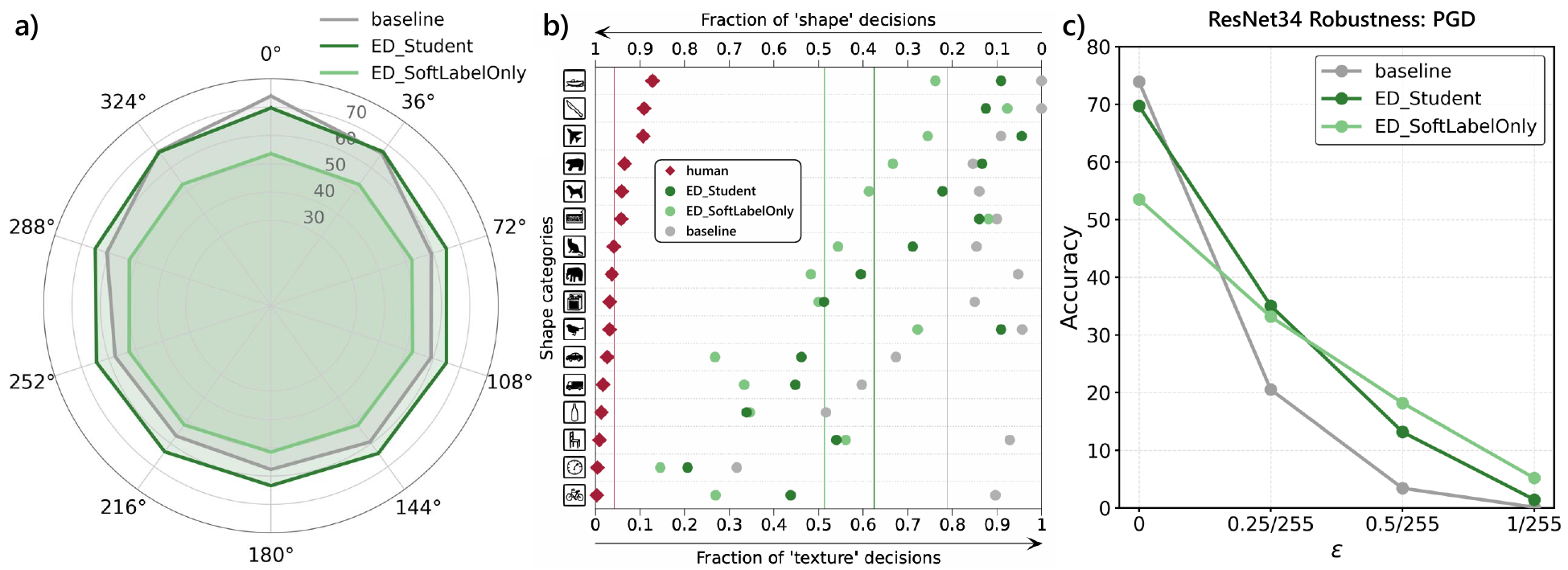}
\vspace{-1.1em}
\caption{
Robustness properties emerging from event-driven knowledge distillation. a) Top-1 accuracy across hue rotation angles. b) Cue-conflict shape bias; Event-distilled students approach the human reference. c) Accuracy under PGD attacks with $0 < \epsilon \le 1/255$.
}
\label{fig:three_phenomenon}
\vspace{-1em}
\end{figure*}

\section{Phenomena: Robustness Properties of Event-Distilled Models}
\label{sec:observations}

We apply the Event-distilled models (ED\_Student, ED\_SoftLabelOnly) to RGB-domain benchmarks to probe their inductive biases. Our experiments show three robustness properties that distinguish these models from conventional CNNs trained only on RGB images. We summarize these main results in Figure~\ref{fig:three_phenomenon} (overview in the right panel of Figure~\ref{fig:method_overview}; comprehensive numbers in Appendices~\ref{sec:ap_color}--\ref{sec:ap_high_freq}).

\if[]
\begin{table}[t]
\centering
\small
\begin{tabular}{lccc}
\toprule
                                         & baseline & ED\_Student & ED\_SoftLabelOnly \\
\midrule
ImageNet val (clean)                      & 73.90 & 69.70 & 53.50 \\
\midrule
\multicolumn{4}{l}{\emph{Color invariance} (Section~\ref{sec:4-1_color})} \\
\quad Hue rotation worst angle            & 56.7  & \textbf{63.4}  & 51.5  \\
\quad Grayscale val                       & 62.50 & \textbf{64.60} & 52.58 \\
\midrule
\multicolumn{4}{l}{\emph{Shape bias} (Section~\ref{sec:4-2_shape})} \\
\quad B\&W val                            & 41.23 & \textbf{48.82} & 34.71 \\
\quad Canny val                           & 10.77 & \textbf{13.23} & 11.38 \\
\bottomrule
\end{tabular}
\caption{Representative top-1 accuracies (\%) for Color invariance and Shape bias. ``Hue rotation worst angle'' reports the minimum accuracy when rotating the hue in $30^\circ$ steps. For high-frequency robustness, see Figure~\ref{fig:three_phenomenon}~(c); for comprehensive numbers, see Appendix~\ref{sec:ap_color}, \ref{sec:ap_shape}, \ref{sec:ap_high_freq}.}
\label{tab:phenomena_summary}
\end{table}
\fi

\subsection{Color and Hue Invariance}
\label{sec:4-1_color}



Standard CNNs are known to rely heavily on object color information for inference~\cite{hosseini2018semantic,engstrom2019spatial,geirhos2019texture_bias}. However, when we evaluate robustness using a dataset whose hue is rotated across the entire image, as shown in Figure~\ref{fig:three_phenomenon}~(a), the baseline model's accuracy drops sharply at certain rotation angles, whereas the Event-distilled model maintains color invariance across all angles. This color invariance is also confirmed when the ImageNet validation set is converted to grayscale (baseline: 73.9$\rightarrow$62.4, ED\_Student: 69.7$\rightarrow$64.6; see Appendix~\ref{sec:ap_color_grayscale} Table~\ref{tab:quantitative_colorless}), suggesting persistence against the loss of color information. Per-model comparisons are reported in Appendix~\ref{sec:ap_color}.

\subsection{High Shape Bias}
\label{sec:4-2_shape}

Next, to clarify what the Event-distilled model relies on for inference if not color, we performed the cue-conflict experiment proposed by Geirhos et al.\ (which classifies images that pit texture and shape information against each other)~\cite{geirhos2019texture_bias}.

The scatter plot in Figure~\ref{fig:three_phenomenon}~(b) shows, for each category, the proportion of model decisions based on texture versus shape. While the baseline model, which performs texture-based inference, lies on the right side of the plot, the Event-distilled models (ED\_Student and ED\_SoftLabelOnly) shift substantially toward the left side and approach the region populated by humans, who make shape-based decisions. In other words, not only do the inference cues switch to shape, but the model also acquires shape-centric decision characteristics similar to those of humans.

This shape bias is also corroborated by patch-shuffle experiments, in which the image is divided into a grid and shuffled~\cite{naseer2021intriguing,brendel2019bagnet}, and by evaluations under conditions in which color and texture information are stripped away to leave only shape (see Appendix~\ref{sec:ap_shape_patchshuffle} Figure~\ref{fig:ap_patch_shuffle} and Appendix~\ref{sec:ap_shape_strip} Table~\ref{tab:quantitative_colorless}). As the mosaic grid becomes finer and the shape in the image is increasingly destroyed, the Event-distilled model's accuracy is observed to drop more sharply than the baseline's.

Two caveats apply to how we read these results. First, Burgert et al.~\cite{burgert2025reliance} show that the preference measured by the cue-conflict experiment does not by itself imply strict reliance, so we describe our result as a shift of preference toward shape rather than a strict reduction of texture reliance. Second, it systematizes controlled-suppression probes, of which our patch-shuffle and patch-rotation protocol is an instance, so we treat the two probes as complementary rather than interchangeable.

\subsection{Robustness to High-Frequency Perturbations}
\label{sec:4-3_highfreq}

The Event-distilled model also acquires resistance to deliberate noise and to adversarial perturbations on the input image.

In general, CNNs use high-frequency texture information as a key cue for classification, which makes them vulnerable to adversarial attacks such as FGSM~\cite{goodfellow2015fgsm} and PGD~\cite{madry2018pgd_adversarialtraining} that target small high-frequency components. As shown by the PGD evaluation in Figure~\ref{fig:three_phenomenon}~(c), the Event-distilled model attains accuracy above the baseline under perturbations with $0 < \epsilon \le 1/255$, and the same trend holds for FGSM. Throughout this paper our adversarial claims are restricted to this perturbation range; we do not evaluate larger budgets or stronger attack ensembles such as AutoAttack~\cite{croce2020autoattack}. Furthermore, in experiments that add band-limited random noise per frequency band, the Event-distilled model is confirmed to be more resistant than the baseline, especially against high-frequency-band noise (see Appendix~\ref{sec:ap_high_freq_fgsm} Figure~\ref{fig:ap_high_freq_robustness} for details of the frequency-band random-noise experiment). PGD evaluation hyperparameters are reported in Appendix~\ref{sec:ap_pgd_eval_details}.

These three phenomena (color invariance, high shape bias, and resistance to high-frequency noise) suggest not a mere accuracy improvement but a shift in the representation space acquired by the model. In the next section, by analyzing the model's internal representation and its responses to real-world artifacts, we elucidate a unified mechanism that explains these seemingly disparate phenomena.

\section{Mechanism: An Edge-Based Shape Feature Hypothesis}
\label{sec:mechanism}

The three phenomena confirmed in the previous section (color invariance, high shape bias, and resistance to high-frequency noise) do not arise independently of one another.
We hypothesize that knowledge distillation from event data causes the student model to suppress its dependence on unstructured high-frequency texture while strengthening its dependence on the macroscopic geometric structure that we call the edge-based shape feature. That is, whereas noise-augmentation methods broadly bolster post-hoc robustness against unstructured perturbations across all frequency bands, our distillation differs in that it selectively enhances dependence on the components that carry edge information. In fact, since event data record only brightness changes, the frequency components that correspond to edges in the original natural image are relatively emphasized, and as a result the radial PSD of N-ImageNet exhibits a smaller power-law exponent ($\alpha = 1.48$) than that of natural images (ImageNet, $\alpha = 2.46$); i.e., a flatter power spectral density (PSD) (see Appendix~\ref{sec:ap_psd} for details).
In this section, we examine this hypothesis through analysis of the model's internal representations and its responses to digital artifacts with different frequency characteristics.

\begin{figure*}
\vspace{-1.5em}
\centering
\includegraphics[width=\linewidth]{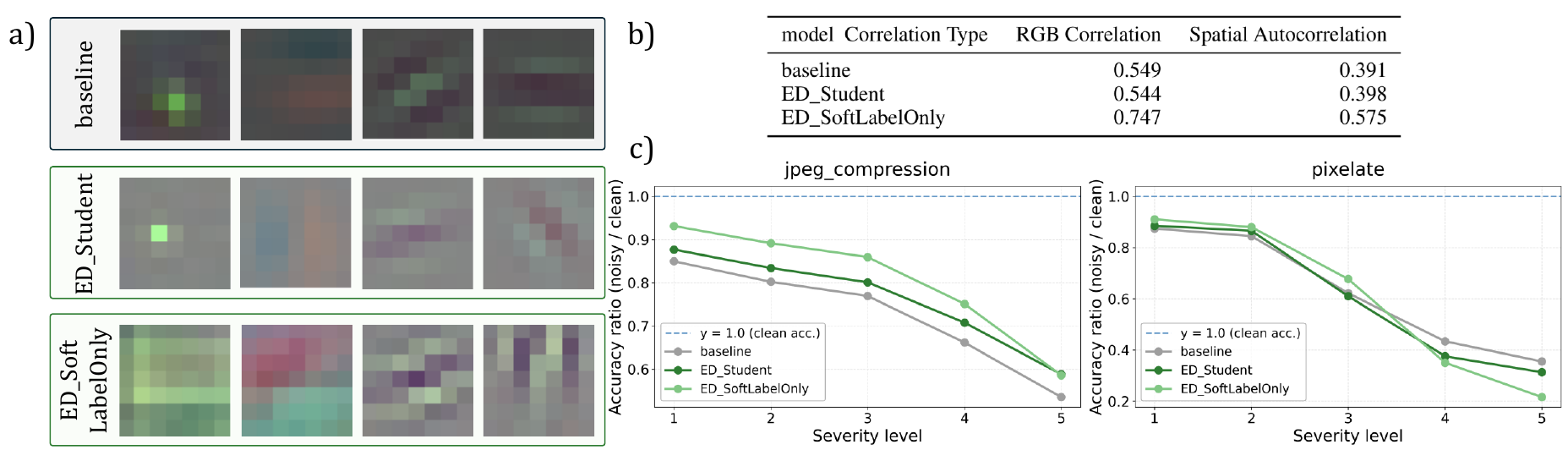}
\vspace{-1.1em}
\caption{
Internal-representation analysis and the spectral trade-off.
a) First-layer filter excerpts, each min-max normalized to $[0,1]$: baseline shows noisy high-frequency patterns; the Event-distilled model (especially ED\_SoftLabelOnly) acquires smooth monochromatic, color-opponent, and Gabor-like structures.
b) RGB and spatial autocorrelations; ED\_SoftLabelOnly exceeds the baseline on both, while ED\_Student stays close to it.
c) Two ImageNet-C corruptions, as accuracy retention (noisy / clean). \emph{Left}: under JPEG compression, the Event-distilled model retains more at high severity. \emph{Right}: under pixelate, a reversal occurs at high severity.
}
\label{fig:filter_jpeg}
\vspace{-1em}
\end{figure*}

\subsection{Internal Representation: Structural Prior in First-Layer Filters}
\label{sec:5-1}

To elucidate the change in the internal representation acquired by the model, we analyzed the input-layer filters (Figure~\ref{fig:filter_jpeg}a). Whereas the baseline exhibits high-frequency noisy patterns that respond to local texture, the Event-distilled model shows high sensitivity to geometric structures. These filters can be broadly classified into (i) monochromatic filters that capture broad color distributions, (ii) color-opponent-like filters that capture macroscopic boundaries, and (iii) directional Gabor-filter-like edge-extraction structures (see Appendix~\ref{sec:ap_shape_filter} for visualization of all 64 channels and detailed functional taxonomy). Importantly, these filters do not merely act as low-pass filters that smooth the input; rather, they acquire a structural prior that suppresses unstructured high-frequency texture while responding selectively to the structural gradients (edges) that constitute object contours. Appendix~\ref{sec:ap_filter_aug_confound} discusses why this is not an augmentation-pipeline artifact.

To quantitatively corroborate this change, we computed the channel-wise (RGB Correlation) and spatial (Spatial Autocorrelation) correlations of the filters (Figure~\ref{fig:filter_jpeg}b). ED\_SoftLabelOnly shows substantially higher correlation values than baseline on both metrics; ED\_Student remains close to baseline; since (b) aggregates over all filters while (a) shows individual ones, the two need not agree (Appendices~\ref{sec:ap_filter_aug_confound} and~\ref{sec:ap_filter_spectra}). This indicates that, in ED\_SoftLabelOnly, all channels synchronously extract geometric structures and the representation space has been transformed to emphasize continuity between adjacent pixels, which underlies the spectral trade-off discussed in the next subsection. How this property operates under actual color changes is corroborated by the early-layer feature-map visualization under hue rotation in Appendix~\ref{sec:ap_color_featmap}.



\subsection{Validation via ImageNet-C: Spectral and Structural Trade-offs}
\label{sec:5-2}

To test the structural prior presented in the previous subsection, we evaluated all 19 corruption types of ImageNet-C. Robustness does not improve uniformly; it exhibits a spectral trade-off that depends on whether the corruption damages the frequency band or the geometric structure.

The Event-distilled model resists perturbations that selectively remove or occlude unstructured high-frequency texture. Under JPEG compression (Figure~\ref{fig:filter_jpeg}c, left) its retention exceeds the baseline's even at high severities, which suggests that stable inference is possible once shortcut texture is lost, as long as the macroscopic brightness gradients (edges) on which the model relies remain.

Conversely, it is vulnerable to perturbations that break the continuity of those geometric structures. Under pixelation (Figure~\ref{fig:filter_jpeg}c, right) it is resistant at low severities, where macroscopic contours are preserved, but at high severities, where edges become discontinuous, a reversal occurs and its retention falls below the baseline. Similar sensitivity appears under contrast and elastic\_transform (all 19 corruptions are analyzed in Appendix~\ref{sec:ap_high_freq_inc}). This trade-off suggests that the decision boundary has shifted from superficial texture toward geometric structure and edge continuity.

\subsection{Validation via Information Removal: Dominance of Geometric Features}
\label{sec:5-3}

To test the edge-based shape feature hypothesis more directly, we evaluate inference under shape-only conditions in which the ImageNet validation set is converted to B\&W (Otsu thresholding~\cite{otsu1979threshold}), Canny edges~\cite{canny1986edge}, or silhouette.
Under these conditions the baseline degrades sharply; ED\_Student also degrades but stays above the baseline on all four (see Appendix~\ref{sec:ap_shape_strip} for the per-model table).

Together with the smooth edge-extracting filters of Section~\ref{sec:5-1}, these results indicate that the inference basis of the Event-distilled model has shifted from texture to geometry (shape).

\section{Validation and Practical Implications}
\label{sec:6_val}

\begin{figure*}[t]
\vspace{-2.2em}
\centering
\includegraphics[width=\linewidth]{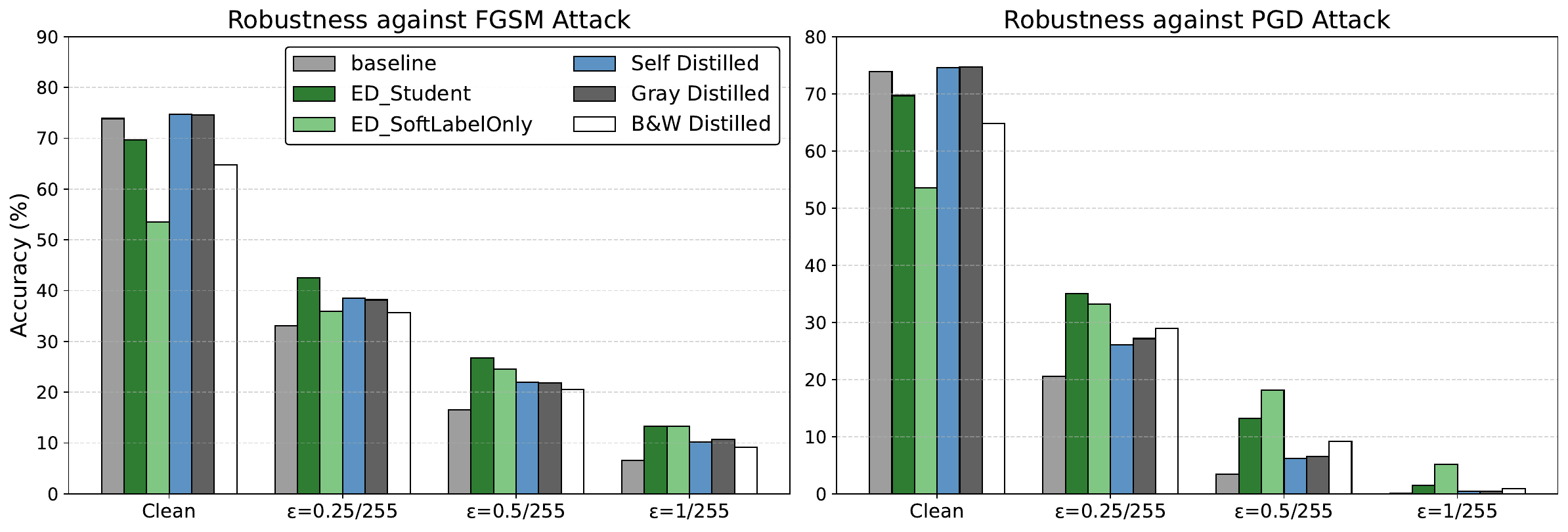}
\vspace{-2.2em}
\caption{
Ablation (Section~\ref{sec:6-1_ablation}). Adversarial robustness of color-stripped controls (Self / Gray / B\&W Distilled) vs.\ the Event-distilled models. \emph{Left}: FGSM. \emph{Right}: PGD. x-axis: perturbation magnitude $\epsilon$; y-axis: top-1 accuracy. The Event-distilled model surpasses all controls at $0 < \epsilon \le 1/255$, indicating its inductive bias is not merely the result of color removal.
}
\label{fig:ablation_adv_attack}
\vspace{-1.5em}
\end{figure*}

The previous sections established that the Event-distilled model acquires an edge-based shape feature. Here we validate it with ablations, compare it against existing robustification techniques, and assess its practical value through Linear-probe transfer on downstream tasks.

\subsection{Ablation Study: Is it merely the absence of color?}
\label{sec:6-1_ablation}

To examine whether the high shape bias and robustness acquired by the Event-distilled model are due to the mere absence of color information, we prepared distillation models from teacher models trained on ImageNet RGB images, grayscale images, and fully color-stripped binary (B\&W) images, and conducted comparative validation against the corresponding distilled students (Self Distilled, Gray Distilled, B\&W Distilled).

Under adversarial attack (Figure~\ref{fig:ablation_adv_attack}), the Event-distilled model resists better than all controls. Its superiority even over B\&W Distilled suggests that the transferred bias is not what one obtains by forcing the student to learn contours, but a representation rooted in sensitivity to the brightness gradients characteristic of the event sensor.

\subsection{Architectural Generality}
\label{sec:6-2_architectural_generality}

\begin{figure*}
\vspace{-1.5em}
\centering
\includegraphics[width=\linewidth]{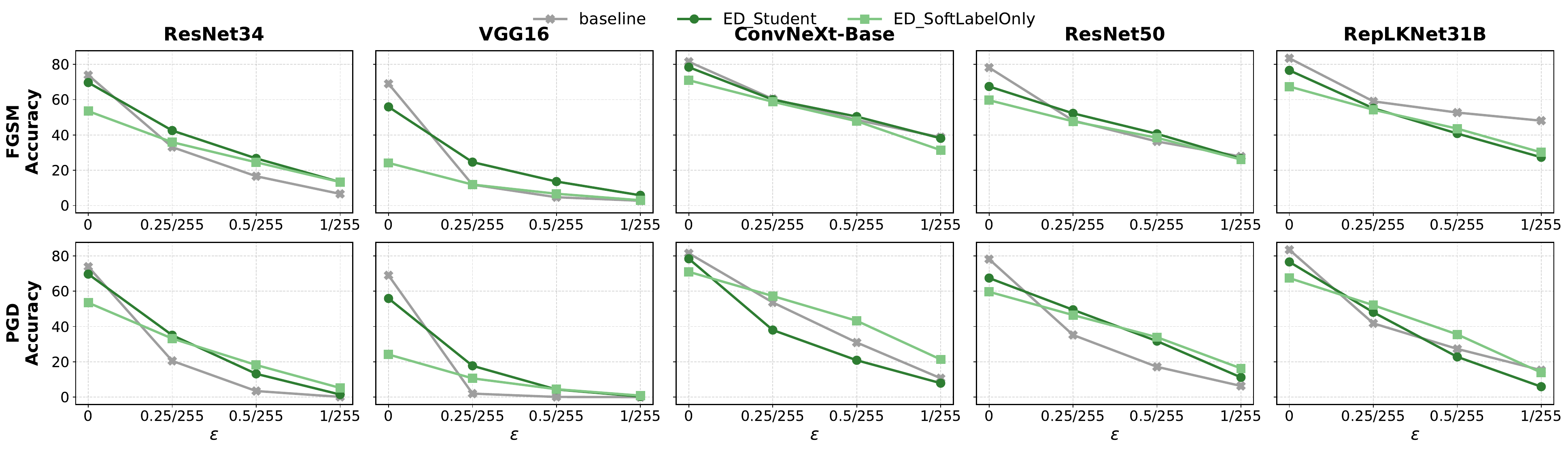}
\vspace{-1.8em}
\caption{
Architectural generality (Section~\ref{sec:6-2_architectural_generality}). Comparison of baseline / ED\_Student / ED\_SoftLabelOnly across ResNet-34, VGG-16, ConvNeXt-Base, ResNet-50, and RepLKNet31B~\cite{ding2022replknet}. Top: FGSM. Bottom: PGD. x-axis: perturbation magnitude $\epsilon$. ED\_Student improves over the baseline in 19 of the 30 tested (architecture, attack, $\epsilon$) settings with $0 < \epsilon \le 1/255$. The 11 exceptions, reported as the ED\_Student\,$-$\,baseline difference in percentage points, are (RepLKNet31B, FGSM) at $\epsilon = 0.25, 0.5, 1$/255: $-3.9$, $-11.9$, $-20.7$; (RepLKNet31B, PGD) at $\epsilon = 0.5, 1$/255: $-4.5$, $-9.4$; (ConvNeXt-Base, PGD) at $\epsilon = 0.25, 0.5, 1$/255: $-15.6$, $-10.1$, $-2.8$; (ConvNeXt-Base, FGSM) at $\epsilon = 0.25, 1$/255: $-0.4$, $-0.6$; and (ResNet-50, FGSM) at $\epsilon = 1$/255: $-1.2$.
}
\label{fig:across_model_adv}
\vspace{-1.5em}
\end{figure*}

Next, we verified whether the phenomena reported in the main paper depend on a particular network structure. In addition to the ResNet-34-based validation, we performed the same Event distillation and evaluation using VGG-16 (a different structure), the deeper ResNet-50, the more recent ConvNeXt-Base, and RepLKNet31B~\cite{ding2022replknet}, a CNN with $31\times 31$ large kernels, and examined two axes (high-frequency noise robustness, Section~\ref{sec:4-3_highfreq}; and shape bias, Section~\ref{sec:4-2_shape}) across these architectures.

\textbf{High-frequency noise robustness.} As shown in Figure~\ref{fig:across_model_adv}, ED\_Student improves over the baseline in 19 of the 30 tested (architecture, attack, $\epsilon$) settings with $0 < \epsilon \le 1/255$ under FGSM and PGD (5 architectures $\times$ 2 attacks $\times$ 3 values of $\epsilon \in \{0.25, 0.5, 1\}/255$). The 11 exceptions are enumerated in the caption of Figure~\ref{fig:across_model_adv}; they are concentrated in the two largest backbones, ConvNeXt-Base and RepLKNet31B.

\textbf{Shape bias.} Both the cue-conflict experiment (Appendix~\ref{sec:ap_shape_cueconflict}) and the patch-shuffle / patch-rotation experiment that destroys macroscopic shape continuity (Appendix~\ref{sec:ap_shape_patchshuffle}) consistently reproduce, across all architectures, the behavior observed for ResNet-34: a shift toward the shape side and a steeper accuracy drop as the grid becomes finer.

These cross-model evaluations suggest that the cross-domain distillation we propose injects its inductive bias without depending on a particular network structure, and that the main benefits reported in the paper, shape bias and high-frequency noise robustness, are largely insensitive to architectural design choices, with the exceptions noted above.

\subsection{Transferability: Event Distillation as a Structural Prior}
\label{sec:6-3_downstream}

An additional finding regarding application is that the inductive bias acquired by the Event-distilled model functions as a useful prior in transfer learning to downstream tasks (Linear probe~\cite{alain2017probes}).
We perform Linear-probe evaluations on 19 image-classification datasets to compare the transfer performance of ED\_Student against baseline. As points of comparison, we prepare representative robustification techniques (Adversarial Training, PGD-based~\cite{madry2018pgd_adversarialtraining}, and Noise Training, Gaussian noise injection~\cite{yin2020noisetraining}) as well as the SIN+IN co-training model (the SIN model)~\cite{geirhos2019texture_bias} that directly induces shape bias, and report the performance gain over the baseline ($\Delta$ Score) in Table~\ref{tab:downstream_task}.
Adversarial Training and Noise Training are reported to enhance robustness against adversarial noise and blur while degrading markedly under perturbations such as fog~\cite{yin2020noisetraining}, in contrast to the Event-distilled model.

Notably, although the Event-distilled model is 4.2 percentage points below the baseline on pre-training ImageNet accuracy, it outperforms the baseline on many downstream tasks under Linear-probe transfer. The largest gains appear on CIFAR-100 / CIFAR-10~\cite{krizhevsky2009cifar}, Chest X-ray~\cite{kermany2018chestxray} and MNIST~\cite{lecun1998mnist}, and gains are also observed where texture is largely absent or fine-grained geometry carries the signal, such as imagenet-sketch~\cite{wang2019sketch} and cars196~\cite{krause2013cars}; per-dataset values are in Table~\ref{tab:downstream_task}. We attribute this to the generality of the representation space acquired during pretraining: by suppressing dataset-specific high-frequency texture and selectively extracting geometric structure, the model acts as a structural prior for unseen tasks under the frozen-weights Linear-probe setting. Hyperparameters and protocol details are in Appendix~\ref{sec:ap_linear_probe_details}.


\begin{table}[t]
\centering
\scriptsize
\setlength{\tabcolsep}{0.8pt}
\renewcommand{\arraystretch}{0.98}
\begin{tabular}{l*{19}{c}|c|c}
\toprule
& \rotatebox{90}{C100} & \rotatebox{90}{C10} & \rotatebox{90}{ChXR} & \rotatebox{90}{MNIST}
 & \rotatebox{90}{Cars} & \rotatebox{90}{DTD} & \rotatebox{90}{STL10} & \rotatebox{90}{INet-sk}
 & \rotatebox{90}{Cal101} & \rotatebox{90}{INet-v2} & \rotatebox{90}{Scene} & \rotatebox{90}{SUN397}
 & \rotatebox{90}{Flwr} & \rotatebox{90}{KITTI} & \rotatebox{90}{ESAT} & \rotatebox{90}{CUB}
 & \rotatebox{90}{PCam} & \rotatebox{90}{Pet} & \rotatebox{90}{Food} & \rotatebox{90}{Mean} & \rotatebox{90}{INet} \\
\midrule
base.\ (\%)
 & 55.8 & 78.0 & 87.3 & 89.9 & 46.9 & 60.1 & 94.8 & 62.2 & 85.4 & 49.6
 & 90.6 & 53.5 & 85.3 & 73.4 & 90.0 & 64.6 & 78.0 & 90.5 & \textbf{61.4} & 73.5 & 73.9 \\
\midrule
ED ($\Delta$)
 & 5.2 & 4.8 & \textbf{4.3} & 3.4 & \textbf{2.3} & \textbf{1.9}
 & \textbf{1.8} & 1.4 & \textbf{1.2} & 0.9 & \textbf{0.8} & \textbf{0.8}
 & \textbf{0.7} & 0.7 & \textbf{0.4} & \textbf{0.3} & 0.2
 & $-$0.1 & $-$0.4 & \textbf{1.6} & $-$4.2 \\
Adv.\ ($\Delta$)
 & \textbf{14.6} & \textbf{10.8} & 2.9 & \textbf{5.6} & $-$8.7 & $-$9.1 & $-$0.7 & $-$5.2 & 0.4 & $-$12.3
 & $-$0.5 & $-$7.8 & $-$9.4 & 2.1 & $-$1.2 & $-$3.4 & 0.0 & $-$3.2 & $-$9.5 & $-$1.8 & $-$7.9 \\
Noise ($\Delta$)
 & 3.2 & 2.6 & 0.6 & $-$0.7 & $-$5.6 & $-$2.5 & 0.6 & $-$4.5 & 0.7 & 1.7
 & $-$0.8 & $-$2.3 & $-$3.9 & 0.4 & $-$1.7 & 0.2 & \textbf{0.3} & $-$0.1 & $-$3.8 & $-$0.8 & \textbf{0.5} \\
SIN ($\Delta$)
 & 6.2 & 5.6 & 0.0 & 1.8 & $-$8.3 & $-$4.2 & 1.4 & $-$4.5 & 0.7 & $-$4.1
 & $-$0.3 & $-$1.9 & $-$5.5 & \textbf{3.2} & $-$2.2 & $-$3.0 & $-$1.2 & $-$0.6 & $-$1.1 & $-$0.9 & $-$1.7 \\
\midrule
Canny ($\Delta$)
 & $-$1.8 & 1.0 & 0.2 & 0.3 & 1.1 & 1.4 & 0.9 & \textbf{1.5} & $-$0.1 & \textbf{2.9}
 & 0.2 & 0.2 & 0.1 & $-$0.1 & $-$1.7 & $-$2.1 & 0.3 & \textbf{0.1} & $-$0.7 & 0.2 & $-$2.8 \\
Synth ($\Delta$)
 & $-$9.4 & $-$6.0 & $-$2.1 & $-$5.0 & $-$4.2 & $-$6.7 & $-$0.1 & $-$4.9 & $-$3.8 & $-$2.4
 & $-$2.3 & $-$11.7 & $-$0.9 & $-$0.9 & $-$7.2 & $-$21.8 & $-$0.6 & $-$1.0 & $-$12.9 & $-$5.4 & $-$2.1 \\
\bottomrule
\end{tabular}
\caption{Linear probe transfer (Sections~\ref{sec:6-3_downstream}, \ref{sec:6-4_vs_robust_methods}). base.\ gives baseline top-1 (\%); the $\Delta$ rows (\%, $+$ omitted) are ED\_Student, Adversarial Training, Noise Training, SIN+IN co-training~\cite{geirhos2019texture_bias}, Canny\_Distilled, and Synth\_Distilled. Mean averages the 19 transfer tasks; INet is the pre-training reference, excluded from Mean. Bold: column-wise best; datasets sorted by ED\_Student's $\Delta$; abbreviations expanded in Appendix~\ref{sec:ap_linear_probe_details}.}
\label{tab:downstream_task}
\vspace{-1.5em}
\end{table}

\subsection{Necessity: What Non-Event Supervisions Fail to Reproduce}
\label{sec:6-4_vs_robust_methods}

\begin{table}[t]
\centering
\scriptsize
\setlength{\tabcolsep}{3pt}
\begin{tabular}{lccccc c}
\toprule
& Clean & $\Delta$ Color (hue Avg.) & $\Delta$ Shape (sketch) & $\Delta$ High-freq (bandpass) & $\Delta$ Transfer (LP Avg.) & Axes improved \\
\midrule
baseline (abs.) & 73.9 & 61.8 & 23.7 & 29.6 & 73.5 & --- \\
\midrule
ED\_Student & 69.7 & $+3.6$ & $+4.4$ & $+17.4$ & $+1.6$ & \textbf{4 / 4} \\
Canny\_Distilled & 71.1 & $+4.3$ & $+4.2$ & $-5.9$ & $+0.2$ & 3 / 4 \\
Synth\_Distilled & 71.8 & $-3.4$ & $-3.2$ & $+18.9$ & $-5.4$ & 1 / 4 \\
Adv.\ Training & 66.0 & $-11.2$ & $-2.4$ & $+28.1$ & $-1.8$ & 1 / 4 \\
Noise Training & 74.4 & $+0.8$ & $+1.7$ & $+13.4$ & $-0.8$ & 3 / 4 \\
SIN & 72.2 & $+0.5$ & $+7.1$ & $+19.3$ & $-0.9$ & 3 / 4 \\
\bottomrule
\end{tabular}
\caption{
Event supervision vs.\ five non-event supervisions, all with the same ResNet-34 student. Clean is absolute for every row; on the four axes the baseline row is absolute and the others are differences from it, signed so that positive means better. Color: mean top-1 over ten hue-rotation angles; Shape: ImageNet-Sketch top-1~\cite{wang2019sketch}; High-freq: reduction in the accuracy drop under band-limited noise at $0.95\pi$, where the baseline drops 29.6\,pp; Transfer: mean linear-probe accuracy over the 19 tasks of Table~\ref{tab:downstream_task}. Only ED\_Student improves on all four axes.
}
\label{tab:joint_profile}
\vspace{-1em}
\end{table}

Is event data required at all for the properties reported above? We compare ED\_Student against five non-event supervisions, all with the same ResNet-34 student, on four axes: the three phenomena of Section~\ref{sec:observations} together with the downstream transfer of Section~\ref{sec:6-3_downstream}. Two of the five are new here: Canny\_Distilled, whose ResNet-34 teacher is trained from scratch on Canny edge maps of ImageNet (teacher 55.82\%, student 71.15\%), representing RGB-derived edge and gradient operators; and Synth\_Distilled, whose teacher is trained on a two-channel ON/OFF proxy formed from the grayscale difference between two augmented views of the same image (teacher 64.52\%, student 71.78\%), representing proxies that imitate event data without an event sensor. The distillation controls follow the ED\_Student recipe; the robustification baselines keep their own (Table~\ref{tab:student-config}). The color-stripped controls of Section~\ref{sec:6-1_ablation} are further non-event supervisions, not repeated here; generation and training details are in Appendix~\ref{sec:ap_added_controls}. Table~\ref{tab:joint_profile} summarizes the resulting four-axis profiles.

Canny\_Distilled reproduces the color and shape aspects: hue rotation (66.1) and ImageNet-Sketch (27.9) are at ED\_Student's level. It nevertheless fails on high-frequency-band robustness, dropping 35.5\,pp at $0.95\pi$ against the baseline's 29.6\,pp, and its downstream gain ($+0.2$\,pp) does not reach ED\_Student's ($+1.6$\,pp). Its cue-conflict shape bias (0.270) also stays closer to the baseline (0.212) than to ED\_Student (0.375): the shape transfer it achieves shows up in sketch-style inputs but not in the shape-versus-texture preference.

Synth\_Distilled shows the complementary pattern: it matches or exceeds ED\_Student on the frequency and adversarial axes (bandpass drop 10.7\,pp; PGD accuracy-curve area 10.2 against 8.9, baseline 6.2), yet falls below the baseline on hue rotation (58.4), ImageNet-Sketch (20.5), and downstream transfer ($-5.4$\,pp). Adversarial Training points the same way: it attains the largest PGD accuracy-curve area (24.1) and the smallest bandpass drop (1.5\,pp) while losing 11\,pp on hue rotation. The adversarial and high-frequency gains are therefore not uniquely attributable to event supervision.

Existing robustification methods are partial in the same way. Adversarial and Noise Training confer targeted resistance to the perturbations they are trained against, but robustness and clean accuracy are known to be at odds~\cite{tsipras2019odds}, and both fall below the baseline in linear-probe transfer. SIN attains the highest accuracy of all models once color and texture are removed (Table~\ref{tab:quantitative_colorless}), yet falls below the baseline on 12 of 19 downstream tasks; since the cue-conflict benchmark is produced by the same style-transfer procedure as Stylized-ImageNet, SIN is evaluated close to its own training distribution and its advantage does not carry over to ordinary RGB images.

Across the six models, ED\_Student is the only one that improves over the baseline on all four axes at once. We therefore do not claim that any single property requires event data; rather, it is the joint profile that the non-event supervisions examined here fail to reproduce.

\textbf{Do the gains survive a stronger baseline?} Retraining both the baseline and ED\_Student under a timm-style recipe (300 epochs with stochastic depth, label smoothing, MixUp, and RandAugment) reaches 75.9\% and 70.8\% clean accuracy. ED's advantage persists on shape ($+2.0$\,pp), color ($+1.8$\,pp), high-frequency-band robustness ($-4.2$\,pp drop), and downstream transfer ($+1.3$\,pp), but not on the adversarial axis, where the stronger augmentation already makes the SOTA baseline robust at these perturbation sizes. Full numbers are in Appendix~\ref{sec:ap_sota_recipe}.

\section{Limitations and Future Work}
\label{sec:limitations}

N-ImageNet is produced by recapturing RGB images on a monitor, so the sensor's high dynamic range is not exercised and the planar source has no motion parallax; our claims are restricted to the brightness-gradient selectivity of event data. The effect also appears to diminish with model capacity: ConvNeXt-Base and RepLKNet31B gain less over their baselines than ResNet-34 and account for most of the 11 exceptions in Figure~\ref{fig:across_model_adv}. Our evaluation is confined to CNNs and image classification. Three directions follow: evaluation on a broader range of vision models beyond CNNs, including Vision Transformers~\cite{dosovitskiy2021vit} and dense-prediction tasks; comparison against feature-level distillation such as FitNets~\cite{romero2015fitnets}, since our loss operates on logits alone; and using Eq.~\ref{eq:eq001} as a fine-tuning objective.

\section{Conclusion}
\label{sec:conclusion}

\if[]
\begin{figure*}
\vspace{-1.5em}
\centering
\includegraphics[width=\linewidth]{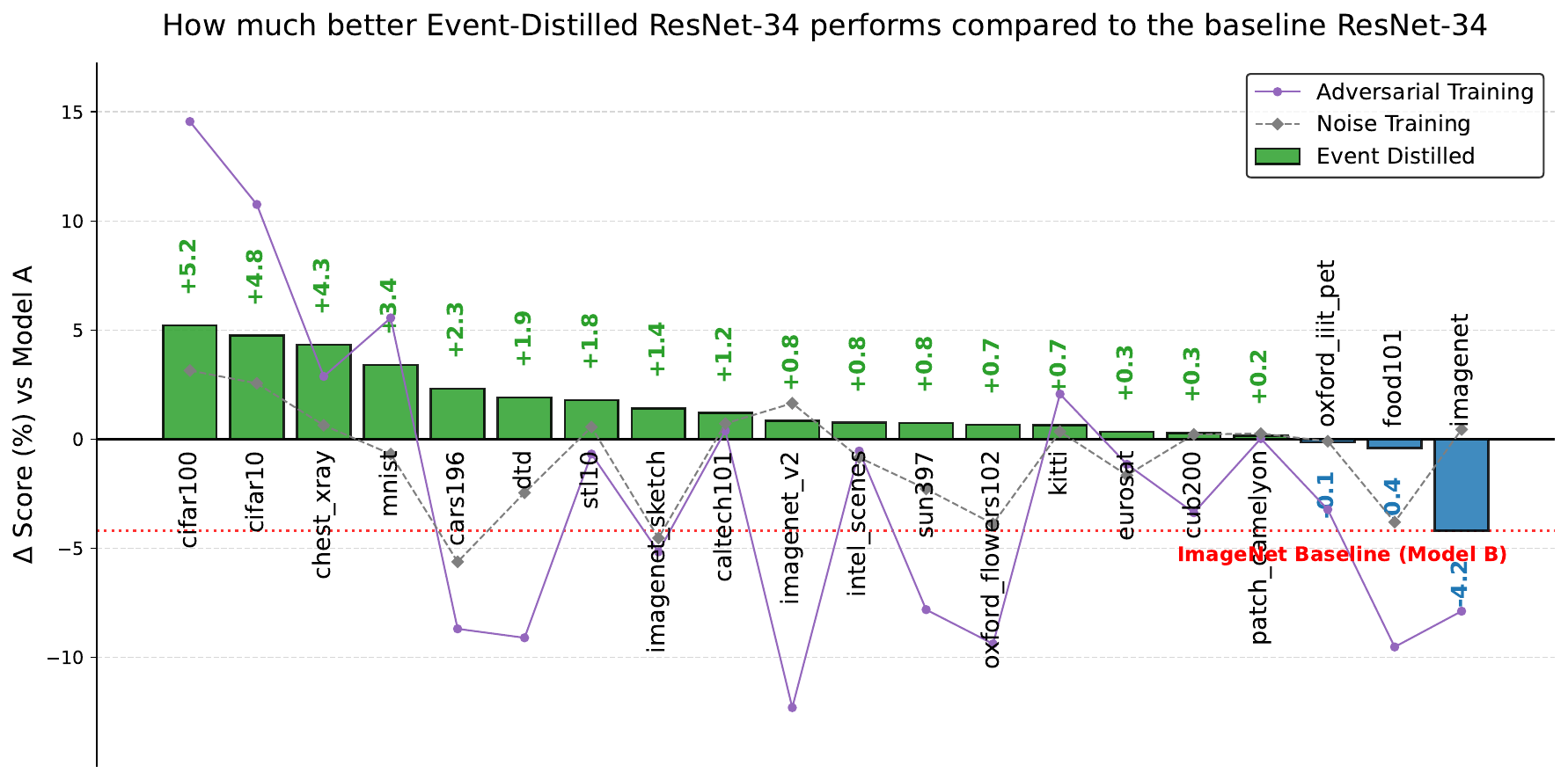}
\vspace{-1.1em}
\caption{
Comparison of transfer performance to downstream tasks (Section~\ref{sec:6-3_downstream}, \ref{sec:6-4_vs_robust_methods}). We report the score difference relative to baseline ($\Delta$ Score, \%) when transferring a pretrained ResNet-34 to 19 image-classification datasets via Linear probe. The green bars indicate the improvement of the Event-distilled model (ED\_Student), the purple line corresponds to Adversarial Training, and the gray line to Noise Training. Datasets are sorted in descending order of the Event-distilled model's $\Delta$ Score. Although the Event-distilled model is inferior to baseline by $-4.2\%$ on clean ImageNet accuracy (rightmost), it clearly surpasses baseline on tasks where shape and structural features matter, such as CIFAR-100, Chest X-ray, and imagenet-sketch --- a clear reversal. By contrast, Adversarial Training and Noise Training cause major performance degradation below baseline on many domains, showing that the inductive bias induced by Event distillation functions as a ``structural prior useful for downstream tasks'' that is distinct from existing robustification techniques.
}
\label{fig:downstream_task}
\vspace{-1em}
\end{figure*}
\fi


To elucidate the inductive bias that event data brings to vision models, we proposed an analysis framework based on cross-domain knowledge distillation. The Event-distilled model suppresses the unstructured high-frequency texture dependence of conventional CNNs and instead depends strongly on the edge-based shape feature. We characterize this as the first systematic empirical study separating what event-domain supervision transfers into RGB representations from what RGB-based alternatives also achieve under a common student architecture.
The mechanism manifests as smoother early-layer filters and a spectral trade-off: robustness to texture loss coexists with vulnerability to frequency-band contamination and geometric disruption. Event data is therefore not only an alternative sensor output but a source of knowledge for more general-purpose visual representations.

\begin{ack}
MT was partially supported by JSPS KAKENHI (22H05116), JST CREST (JPMJCR22N4), and AMED under Grant Number JP25wm0625422.
\end{ack}

{
\small
    \bibliographystyle{unsrtnat}
    \bibliography{main}
}


\appendix

\appendix

\section{Experimental Setup}
\label{sec:ap_exp_setup}

\subsection{Datasets}

In this work, we adopt Neuromorphic-ImageNet (N-ImageNet)~\cite{kim2021nimagenet} as the dataset for the event domain, and ILSVRC2012 (ImageNet) as the dataset for the RGB domain. N-ImageNet has the same 1000-class structure as ImageNet and is generated by displaying each image of ImageNet on a monitor and recapturing it with a DVS-type event camera. As such, every event sample in N-ImageNet is paired at the instance level with the corresponding RGB image in ImageNet. The Stylized-ImageNet dataset~\cite{geirhos2019texture_bias}, used in our SIN comparison (Section~\ref{sec:6-4_vs_robust_methods}), is likewise a 1000-class style-transferred variant of ImageNet with the same 1{,}281{,}167 training images.

The splits used for training and evaluation are as follows.

\begin{itemize}
    \item Teacher training (Section A.2): Only the training split of N-ImageNet is used for training, and the validation split is used for verification.
    \item Student training (Section A.3): Strongly augmented RGB inputs from the ImageNet training split are used to compute the CE loss, while the events from the N-ImageNet training split corresponding to the same instance are passed through the teacher model to compute the KL loss. Validation is performed on the ImageNet validation split with top-1/top-5 accuracy.
\end{itemize}

\subsection{Teacher Training}
\subsubsection{Event Teacher (DiST ResNet-34)}
\label{sec:ap_a21}

As the Event teacher in this work, we directly use the DiST ResNet-34 checkpoint released by the authors of the prior work~\cite{kim2021nimagenet}, in a frozen state. This model is a ResNet-34 that takes the DiST representation as input, and it achieves 48.43\% top-1 accuracy on the N-ImageNet validation split.

In all Event-distillation experiments with ResNet-34 (ED\_Student, ED\_SoftLabelOnly, and the transfer experiments from Section~\ref{sec:6-3_downstream} onward), we use these frozen weights as a common teacher.

\subsubsection{Teacher Models for Comparison}
\label{sec:ap_a22}

The control teachers used in the ablation of Section~\ref{sec:6-1_ablation} are all ResNet-34 variants trained on ImageNet with the CE loss.

\begin{itemize}
    \item Teacher for Self Distilled: A ResNet-34 baseline trained normally on RGB ImageNet is used as the teacher.
    \item Teacher for Gray Distilled: A ResNet-34 trained on the ImageNet RGB images converted to 3-channel grayscale.
    \item Teacher for B\&W Distilled: A ResNet-34 trained on ImageNet preprocessed by OpenCV's Otsu thresholding~\cite{otsu1979threshold} (cv2.THRESH\_BINARY | cv2.THRESH\_OTSU). To preserve the binary property, all spatial transforms use NEAREST interpolation (see Section~\ref{sec:aug-bw-pipe}).
\end{itemize}

These three teachers are all trained for 120 epochs using AdamW~\cite{loshchilov2019adamw} (learning rate $= 2 \times 10^{-3}$, weight decay $= 5 \times 10^{-2}$), with 5 epochs of linear warmup followed by cosine scheduling~\cite{loshchilov2017sgdr} that decays to 0 over the remaining epochs. CutMix and MixUp are not applied.

\subsubsection{Additional Teacher Models for Architectural Generality}
\label{sec:ap_a23}

In the architectural-generality validation of Section~\ref{sec:6-2_architectural_generality}, we additionally trained DiST teachers in-house for 4 architectures other than ResNet-34 (ResNet-50, VGG16-BN, ConvNeXt-Base, RepLKNet31B).
Note that the DiST ResNet-34 teacher~\cite{kim2021nimagenet} of Section~\ref{sec:ap_a21} uses the publicly released weights as-is, and is therefore not part of the additional training in Section~\ref{sec:6-2_architectural_generality}.

The training recipe shared across the four architectures (input representation, event slicing, batch size, weight decay, momentum, number of epochs, parallelization scheme) is shown in Table~\ref{tab:teacher-recipe}, while the architecture-specific settings (optimizer, learning rate, first-conv kernel size) are shown in Table~\ref{tab:teacher-section62}. CutMix and MixUp are enabled only when training the RepLKNet31B teacher (see Section~\ref{sec:aug-cutmix-mixup}).

\begin{table}[h]
\centering
\begin{tabular}{ll}
\toprule
Item & Value \\
\midrule
Input representation & DiST (\texttt{reshape\_then\_acc\_adj\_sort}), 2\,ch, $224 \times 224$ \\
Event slicing        & Random, \texttt{slice\_length} $= 30{,}000$ \\
Batch size           & 512 \\
Weight decay         & $1 \times 10^{-4}$ \\
Momentum             & 0.9 \\
Epochs               & 100 \\
Parallelism          & DataParallel \\
\bottomrule
\end{tabular}
\caption{Shared training recipe for the in-house DiST teachers of Section~\ref{sec:6-2_architectural_generality}. Architecture-specific settings are in Table~\ref{tab:teacher-section62}. The DiST ResNet-34 teacher used elsewhere is the publicly released checkpoint of~\cite{kim2021nimagenet} and is not retrained by us.}
\label{tab:teacher-recipe}
\end{table}

\begin{table}[h]
\centering
\begin{tabular}{llcc}
\toprule
Architecture & Optimizer & Learning rate & First-conv kernel size \\
\midrule
ResNet-50     & Adam  & $2 \times 10^{-4}$ & 14 \\
VGG16-BN      & Adam  & $2 \times 10^{-4}$ & 14 \\
ConvNeXt-Base & AdamW & $2 \times 10^{-4}$ & 4  \\
RepLKNet31B   & AdamW & $2 \times 10^{-4}$ & 4  \\
\bottomrule
\end{tabular}
\caption{Architecture-specific settings for the DiST teachers in Section~\ref{sec:6-2_architectural_generality}. Common settings are in Table~\ref{tab:teacher-recipe}. The DiST ResNet-34 teacher uses the public checkpoint of~\cite{kim2021nimagenet} and is omitted.}
\label{tab:teacher-section62}
\end{table}

\subsection{Student Training}
This subsection describes in detail the training settings for the student models in the main paper. All settings below pertain to the ResNet-34 student.

\subsubsection{Loss Function}

All KD students minimize the composite loss given in Eq.~\ref{eq:eq001} of the main paper. The mixing weight $\alpha$ corresponding to each model and the teacher used are shown in Table~\ref{tab:student-losses}. The temperature in $\mathcal{L}_{\mathrm{KL}}$ is fixed at $T = 3.0$ across all KD students.

For the Adversarial Training and Noise Training models, the KL term is not used (corresponding to $\alpha = 0$); only a cross-entropy loss with label smoothing $\varepsilon = 0.1$ is minimized. The SIN model is trained on the union of ImageNet and Stylized-ImageNet with a plain (non-smoothed) cross-entropy loss only (see Section~\ref{sec:ap_a34}).

\begin{table}[h]
\centering
\begin{tabular}{lcl}
\toprule
Model & $\alpha$ & Teacher \\
\midrule
baseline                      & 0.0 & -- \\
\textbf{ED\_Student} (main)   & 0.2 & DiST ResNet-34 \\
ED\_SoftLabelOnly             & 1.0 & DiST ResNet-34 \\
\midrule
Self Distilled                & 0.2 & RGB ResNet-34 baseline \\
Gray Distilled                & 0.2 & Grayscale ResNet-34 \\
B\&W Distilled                & 0.2 & B\&W ResNet-34 \\
\bottomrule
\end{tabular}
\caption{Distillation weight $\alpha$ (Eq.~\ref{eq:eq001}) and the teacher used for each ResNet-34 student. The KL-term temperature is fixed at $T = 3.0$. Adversarial / Noise Training and the SIN model use only a CE loss and are omitted from this table (see Section~\ref{sec:ap_a33}, \ref{sec:ap_a34}, respectively).}
\label{tab:student-losses}
\end{table}

\subsubsection{Optimization and Learning Rate Scheduling}

For all ResNet-34 student models, we adopt a batch size of 512, AdamW as the optimizer, a common random seed (seed=1), and DDP parallelization. Per-model learning rates, numbers of epochs, and learning-rate schedules are shown in Table~\ref{tab:student-config}.

\begin{table}[h]
\scriptsize
\centering
\begin{tabular}{lcccl}
\toprule
Model & LR & Weight decay & Epochs & LR schedule \\
\midrule
baseline / ED\_Student / ED\_SoftLabelOnly & $2 \times 10^{-3}$   & $5 \times 10^{-2}$ & 120 & 5-epoch linear warmup $\rightarrow$ cosine \\
Self / Gray / B\&W Distilled                & $7.5 \times 10^{-4}$ & $5 \times 10^{-2}$ & 120 & 5-epoch linear warmup $\rightarrow$ cosine \\
SIN                                          & $2 \times 10^{-3}$   & $5 \times 10^{-2}$ & 120 & 5-epoch linear warmup $\rightarrow$ cosine \\
\midrule
Adversarial Training                        & $1 \times 10^{-3}$   & $2 \times 10^{-4}$ & 50  & Cosine annealing (no warmup) \\
Noise Training                              & $1 \times 10^{-3}$   & $2 \times 10^{-4}$ & 120 & Cosine annealing (no warmup) \\
\bottomrule
\end{tabular}
\caption{Per-student optimization configuration for the ResNet-34 student models.}
\label{tab:student-config}
\end{table}

\subsubsection{Additional Settings for Adversarial Learning/Noise Learning}
\label{sec:ap_a33}

\begin{itemize}
    \item PGD attack (Adversarial Training): perturbation bound $\epsilon = 8/255$, step size $\alpha = 2/255$, number of iterations $K = 10$. For each batch, after initializing with uniform random noise of magnitude $\epsilon$, we perform $K$ signed-gradient updates and clip back to within $\epsilon$ in the $L_\infty$ norm at the end of each step. The student is trained only on the generated adversarial inputs, and validation is performed on clean images.
    \item Gaussian Noise Injection (Noise Training): Gaussian noise with mean 0 and standard deviation $\sigma = 0.1$ is added to the tensor image with probability $p = 0.5$, and the result is clipped to $[0, 1]$. This layer is inserted immediately before Normalize. No noise is added at validation.
\end{itemize}

\subsubsection{Additional Settings for SIN+IN Co-training}
\label{sec:ap_a34}

Following the SIN+IN variant of \citet{geirhos2019texture_bias}, the SIN model is trained as a single ResNet-34 on a ConcatDataset that joins the training splits of ImageNet and Stylized-ImageNet (2{,}562{,}334 images in total). Each epoch draws samples from the two datasets in a 1:1 ratio. Both datasets share the same 1000-class WNID set as ImageNet, and we assert in the loader that the two share the same $class\_to\_idx$. Validation is performed on the ImageNet validation split.

The KD term is not used and label smoothing is not applied; only a plain cross-entropy loss is minimized. The optimizer, schedule, seed, and parallelization are kept identical to baseline / ED\_Student (Table~\ref{tab:student-config}), so that no confound other than the dataset is introduced. The augmentation pipeline is also the same RGB strong pipeline (Table~\ref{tab:aug-rgb-strong}) applied identically to both datasets. Trained for 120 epochs under this setup, the model achieves a top-1 accuracy of 72.18\% on the ImageNet validation split.

\subsection{Data Augmentation}
\label{sec:aug}

\subsubsection{RGB pipelines (students and RGB-based teachers)}
\label{sec:aug-rgb}

We use two RGB augmentation pipelines, chosen by purpose:
\begin{itemize}
    \item a \textbf{strong} pipeline on the CE branch of every student
          (Table~\ref{tab:aug-rgb-strong}), and
    \item a \textbf{weak} pipeline on the KD branch and on every validation
          loader (Table~\ref{tab:aug-rgb-weak}).
\end{itemize}

RandAugment is applied to every batch, i.e.\ with probability $p = 1.0$. Adversarial Training uses a simplified strong pipeline that omits RandAugment and RandomErasing, so that the augmentations do not interfere with the PGD perturbations. Noise Training keeps the full strong pipeline and additionally inserts the Gaussian-noise layer described in Section~\ref{sec:ap_a33} before Normalize.

\subsubsection{Grayscale pipelines}
\label{sec:aug-gray}

For the Grayscale teacher and for the KD branch of Gray Distilled, we extend the RGB pipelines of Section~\ref{sec:aug-rgb} by inserting \texttt{torchvision.transforms.v2.Grayscale(num\_output\_channels=3)} (i.e.\ after all spatial and RandAugment operations). The ImageNet normalization statistics (mean and standard deviation) are reused unchanged, applied to the three-times replicated luminance channel.

\subsubsection{B\&W pipelines}
\label{sec:aug-bw-pipe}

Binarized inputs are used by the B\&W teacher and by the KD branch of B\&W Distilled. Preserving the two-valued pixel distribution is critical here, so we (i) use \textsc{Nearest} interpolation for every spatial operation, and (ii) drop the pixel-smoothing augmentations RandAugment and RandomErasing. The resulting pipeline is listed in Table~\ref{tab:aug-bw}.

The B\&W images themselves are produced offline by OpenCV's Otsu thresholding (\texttt{cv2.THRESH\_BINARY | cv2.THRESH\_OTSU}). A fallback path applies the same thresholding on the fly whenever the precomputed image is missing from disk.

\subsubsection{Event-side augmentation}
\label{sec:aug-event}

Event inputs use only three operations: random time flip, random horizontal flip, and random spatial shift (up to $\pm 20$~pixels). No counterpart to the strong RGB pipeline --- i.e.\ color- or texture-based augmentations such as RandAugment or RandomErasing --- is applied on the event side. These three event-side operations are sampled independently per iteration during distillation, matching the recipe used to train the original DiST teacher~\cite{kim2021nimagenet}.

\subsubsection{Cross-Modal Augmentation Alignment in the KD Branch}
\label{sec:ap_kd_view_alignment}

A consequence of the pipelines described above is that, at distillation time, the teacher's event input $x_{\text{event}}$ is passed through the stochastic event-side pipeline of Section~\ref{sec:aug-event} (random time flip, random horizontal flip, $\pm 20$\,px spatial shift), whereas the student's KD branch consumes a deterministic Resize/CenterCrop view $x_{\text{weak}}$ of the paired RGB image (Table~\ref{tab:aug-rgb-weak}). The two views of a given paired instance are therefore not synchronized on a per-iteration basis. We argue below that this asymmetry does not introduce systematic label noise into the distillation target.

\textbf{(A) The teacher--student view asymmetry follows established KD practice.} Asymmetric augmentation between the teacher and student branches is standard in modern distillation. DeiT~\cite{touvron2021deit} feeds the teacher with weakly augmented inputs while applying strong augmentations (RandAugment, MixUp, CutMix) to the student; Patient Teacher~\cite{beyer2022patient} studies teacher--student view consistency in distillation and reports that feeding the teacher and the student the same view is beneficial. Our cross-modal setting cannot share a view, since the two branches consume different modalities; the resulting view mismatch is a confound that we did not isolate experimentally.

\textbf{(B) The KL is computed in class-probability space; spatial misalignment is not pixel-level label noise.} The teacher's event-side augmentations (random horizontal flip, $\pm 20$\,px spatial shift) preserve the class identity of the underlying instance: a flipped or slightly shifted view of a butterfly event is still a butterfly. The KL divergence in Eq.~\ref{eq:eq001} is computed between class-probability distributions $p_t$ and $p_s$~\cite{hinton2015knowledge_distillation}, not between pixel-aligned predictions. Hence the teacher--student view mismatch translates into a stochastic regularization signal --- the student is asked to match the teacher's class distribution across slightly perturbed views of the same instance --- rather than into systematic label noise. This is analogous to consistency regularization in self-supervised learning, where two augmented views of the same image are pulled together in representation space.

\textbf{(C) Empirical coherence of behavioral signatures is inconsistent with systematic noise.} If the cross-modal view asymmetry produced systematic label noise that distorted the distillation target, we would expect ED\_Student / ED\_SoftLabelOnly to exhibit degraded or random behavior. Instead, both models display a coherent and trackable set of signatures aligned with the edge-based shape feature hypothesis: color invariance (Section~\ref{sec:4-1_color}), shape bias (Section~\ref{sec:4-2_shape}), high-frequency-noise robustness (Section~\ref{sec:4-3_highfreq}), the JPEG / pixelate trade-off (Section~\ref{sec:5-2}), shape-only inference (Section~\ref{sec:5-3}), and downstream transfer (Section~\ref{sec:6-3_downstream}). Stochastic label noise would not yield such a directional and reproducible representation shift; the empirical evidence is therefore consistent with the view-mismatch acting as a benign regularizer rather than as a source of systematic distortion.

\subsubsection{Scope of CutMix and MixUp}
\label{sec:aug-cutmix-mixup}

CutMix~\cite{yun2019cutmix} and MixUp~\cite{zhang2018mixup} are used only when training the RepLKNet31B teacher and its corresponding student. When both are enabled, one of the two is selected uniformly at random ($p = 0.5$) per mini-batch.
For every other model --- the ResNet-34, ResNet-50, VGG16-BN, and ConvNeXt-Base teachers, as well as every ResNet-34 student studied in this paper --- neither CutMix nor MixUp is applied, except for the stronger-recipe runs of Appendix~\ref{sec:ap_sota_recipe}, which use MixUp.

\begin{table}[h]
\centering
\begin{tabular}{ll}
\toprule
Operation & Setting \\
\midrule
RandomResizedCrop      & output $224$, scale $= (0.08, 1.0)$ \\
RandomHorizontalFlip   & $p = 0.5$ \\
RandAugment            & \texttt{num\_ops} $= 2$, \texttt{magnitude} $= 9$ \\
Normalize              & ImageNet mean/std \\
RandomErasing          & $p = 0.25$ \\
\bottomrule
\end{tabular}
\caption{RGB strong augmentation pipeline used on the CE branch of every ResNet-34 student. Adversarial Training omits RandAugment~\cite{cubuk2020randaugment} and RandomErasing~\cite{zhong2020random_erasing} to avoid interference with PGD noise.}
\label{tab:aug-rgb-strong}
\end{table}

\begin{table}[h]
\centering
\begin{tabular}{ll}
\toprule
Operation & Setting \\
\midrule
Resize      & 256 \\
CenterCrop  & 224 \\
Normalize   & ImageNet mean/std \\
\bottomrule
\end{tabular}
\caption{RGB weak augmentation pipeline used on the KD branch (student input during KD loss computation) and on every validation loader.}
\label{tab:aug-rgb-weak}
\end{table}

\begin{table}[h]
\centering
\begin{tabular}{ll}
\toprule
Operation & Setting \\
\midrule
RandomResizedCrop      & output $224$, scale $= (0.08, 1.0)$, \textbf{NEAREST} interpolation \\
RandomHorizontalFlip   & $p = 0.5$ \\
Normalize              & ImageNet mean/std \\
\bottomrule
\end{tabular}
\caption{Augmentation pipeline for binarized (B\&W) inputs. All spatial operations use NEAREST interpolation to preserve the two-valued pixel distribution; RandAugment and RandomErasing are omitted.}
\label{tab:aug-bw}
\end{table}

\subsection{Downstream Linear-Probe Protocol and Hyperparameters}
\label{sec:ap_linear_probe_details}

The downstream linear-probe evaluation reported in Section~\ref{sec:6-3_downstream} of the main paper follows a strict linear-probe protocol: the entire ResNet-34 backbone is frozen, and only a single newly initialized linear layer (\texttt{nn.Linear(in\_features, num\_classes)}) is trained on each downstream dataset. This isolates the transferability of the pretrained features and prevents the backbone capacity from confounding the comparison between models.

Training hyperparameters are kept identical across all five compared models (baseline, ED\_Student, Adversarial Training, Noise Training, SIN) and across all 19 downstream datasets:
\begin{itemize}
    \item Optimizer: SGD with momentum $0.9$, weight decay $1{\times}10^{-4}$.
    \item Learning rate: $0.1$ with cosine annealing (\texttt{CosineAnnealingLR}, $T_{\max}=$ epochs).
    \item Epochs: $50$, batch size: $1024$, loss: cross-entropy without label smoothing.
    \item Train transform: \texttt{RandomResizedCrop}(224, scale $0.08$--$1.0$) + \texttt{RandomHorizontalFlip} + \texttt{RandAugment}(\texttt{num\_ops}$=2$, \texttt{magnitude}$=9$) + ImageNet \texttt{Normalize} + \texttt{RandomErasing}($p=0.25$).
    \item Validation transform: \texttt{Resize}(256, antialias) + \texttt{CenterCrop}(224) + ImageNet \texttt{Normalize}.
    \item Two seeds per (model, dataset) pair; the higher of the two best validation accuracies is reported in Table~\ref{tab:downstream_task}.
\end{itemize}

The column abbreviations used in Table~\ref{tab:downstream_task} are C100=CIFAR-100, C10=CIFAR-10, ChXR=Chest X-ray, MNIST=MNIST, Cars=cars196, DTD=dtd, STL10=stl10, INet-sk=imagenet-sketch, Cal101=caltech101, INet-v2=imagenet-v2, Scene=intel\_scenes, SUN397=sun397, Flwr=oxford\_flowers102, KITTI=kitti, ESAT=eurosat, CUB=cub200, PCam=patch\_camelyon, Pet=oxford\_iiit\_pet, Food=food101, and INet=ImageNet (the pre-training reference).

A note on absolute baselines. Strict linear probing of frozen ImageNet-pretrained ResNet-34 features on small-image datasets such as CIFAR-100 (32$\times$32 native resolution upsampled to 224$\times$224) is known to yield substantially lower numbers than (i) full backbone fine-tuning, (ii) larger backbones, or (iii) dedicated low-resolution recipes. The baseline accuracies in Table~\ref{tab:downstream_task} (e.g., $55.8\%$ on CIFAR-100, $78.0\%$ on CIFAR-10) are therefore consistent with the chosen protocol, which deliberately freezes the backbone to measure the transferability of the pretrained representation rather than the joint fine-tuning capacity. We use the same protocol for all compared models, so the relative differences ($\Delta$ Score in Table~\ref{tab:downstream_task}) remain meaningful even when the absolute baseline numbers are protocol-bound.

\subsection{Compute and Runtime}
\label{sec:compute}

All training in this work was conducted on a single server equipped with $8\times$ NVIDIA A100 GPUs. The software stack was Python~3.10, PyTorch~2.1.0, and CUDA~12.2.

Reproducing the models of this paper takes 9 variants over 11 runs, about 11 days of wall-clock time on this hardware: five distillation runs at roughly 1.5 days each (ED\_Student, ED\_SoftLabelOnly, and the Self / Gray / B\&W Distilled variants) and six standard runs at roughly 0.6 days each (baseline, the Gray and B\&W teachers, Adversarial Training, Noise Training, and SIN). The event teacher requires no training, since we use the public DiST checkpoint~\cite{kim2021nimagenet}. Verifying the central claim needs only two of these runs, the baseline and ED\_Student, or about two days. To make cross-checking cheaper we release the code, all trained checkpoints so that every evaluation in this paper can be reproduced without retraining, and an ImageNet-100 configuration for lightweight replication.

\subsection{Summary of Trained Models and Clean Validation Accuracy}
\label{sec:summary}

Table~\ref{tab:clean-accuracy} consolidates, in one place, the clean top-1 validation accuracy of the teacher and student models used in the main comparisons. The DiST column reports each teacher's top-1 accuracy on the N-ImageNet validation split (event domain); the remaining three columns report the corresponding student's top-1 accuracy on the ImageNet validation split (RGB domain).

Two points are worth noting:
\begin{itemize}
\item The DiST ResNet-34 row (48.43\%) corresponds to the publicly
      released checkpoint of~\cite{kim2021nimagenet} that we reuse as-is
      (Section~\ref{sec:ap_a21}); the remaining DiST rows are teachers we trained
      in-house following the recipe of Section~\ref{sec:ap_a23}.
\item ED\_SoftLabelOnly exhibits a wide spread (24.14--70.97\%) because
      its student receives only the teacher's soft labels ($\alpha = 1$).
      When the teacher is strong relative to the student backbone
      (ConvNeXt-Base, RepLKNet31B), the signal transfers effectively; a
      weaker teacher paired with a weaker backbone (VGG16-BN) instead
      collapses the student. ED\_Student improves on ED\_SoftLabelOnly for every
      architecture but stays 3.10--13.11 percentage points below the
      corresponding baseline, consistent with the
      clean-accuracy / robustness trade-off reported throughout the main
      paper.
\end{itemize}


\subsubsection{Teacher Confidence and the Ablation Confounder}
\label{sec:ap_teacher_confidence}

The Event teacher (DiST ResNet-34) attains 48.43\% top-1 on N-ImageNet, which is substantially lower than the ImageNet-1k accuracy of an RGB-trained ResNet-34 ($\approx 73.9\%$). One could therefore worry that the differing soft-label entropy of the teachers used in the ablation study (Section~\ref{sec:6-1_ablation}) confounds the modality comparison: the ED\_Student behavior might reflect high-entropy KL targets rather than event-modality content. We argue this confound is bounded for three reasons.

\textbf{(A) The control teachers are not uniformly highly confident.} The Self Distilled teacher receives standard RGB inputs and converges close to baseline accuracy, but the Gray Distilled and B\&W Distilled teachers are fed luminance-only and binarized inputs respectively at distillation time; these inputs carry strictly less information than the original RGB, so the per-input confidence of these control teachers is expected to be lower than that of the Self Distilled teacher. The asymmetry is therefore not a strict low-entropy-controls vs.\ high-entropy-DiST dichotomy.

\textbf{(B) Entropy alone does not produce directional behavioral signatures.} High-entropy soft labels --- e.g., from temperature-scaled distillation~\cite{hinton2015knowledge_distillation} or label smoothing --- can affect calibration and slightly aid generalization, but they are not known to selectively produce edge-tuned first-layer filters, cue-conflict shape bias, JPEG / pixelate trade-offs, or transfer gains on shape-relevant downstream tasks. The directional pattern observed for ED\_Student matches the edge-based shape feature hypothesis, not a generic high-entropy regularization effect.

\textbf{(C) ED\_SoftLabelOnly is difficult to reconcile with the entropy-driven explanation.} If the high-entropy DiST KL targets diluted the supervision, then ED\_SoftLabelOnly ($\alpha = 1.0$, trained entirely via the high-entropy KL) should show weaker event-domain signatures than ED\_Student. Empirically, ED\_SoftLabelOnly exhibits the strongest signatures (highest filter correlation in Table~\ref{tab:filter_correlation}, most pronounced shape bias in Appendix~\ref{sec:ap_shape_cueconflict}), the opposite of what an entropy-confound explanation would predict.

\begin{table}[h]
\centering
\begin{tabular}{lcccc}
\toprule
              & Teacher & \multicolumn{3}{c}{Student (ImageNet)} \\
\cmidrule(lr){2-2} \cmidrule(lr){3-5}
Architecture  & DiST    & ED\_SoftLabelOnly & ED\_Student & baseline \\
\midrule
ResNet-34     & 48.43 & 53.50 & 69.70 & 73.90 \\
VGG16-BN      & 42.13 & 24.14 & 55.90 & 69.01 \\
ConvNeXt-Base & 52.42 & 70.97 & 78.40 & 81.50 \\
ResNet-50     & 51.11 & 59.68 & 67.47 & 78.16 \\
RepLKNet31B   & 51.61 & 67.43 & 76.61 & 83.45 \\
\bottomrule
\end{tabular}
\caption{Clean top-1 accuracy (\%) of the teacher and student models used in the main comparisons. The Teacher column (DiST) is evaluated on N-ImageNet validation (event domain); the Student columns are on ImageNet validation (RGB domain). The two additional non-event controls are listed separately in Table~\ref{tab:added-controls}.}
\label{tab:clean-accuracy}
\end{table}

\subsection{Additional Non-Event Controls}
\label{sec:ap_added_controls}

Section~\ref{sec:6-4_vs_robust_methods} uses two further non-event supervisions. Both keep the architecture and the distillation recipe of ED\_Student fixed, so that the teacher's input modality is the only experimental variable.

\paragraph{Canny\_Distilled (RGB-derived edge teacher).}
Each RGB ImageNet image is converted to a binary edge map with \texttt{cv2.Canny}(\texttt{threshold1}$=100$, \texttt{threshold2}$=200$) and the map is replicated to three channels so that the input shape matches RGB. The teacher is a standard torchvision ResNet-34 trained from scratch on these edge maps; apart from the Canny thresholds, its recipe is identical to the comparison teachers of Section~\ref{sec:ap_a22} (AdamW, bf16-mixed, 120 epochs, effective batch 512). We chose Canny because \texttt{cv2.Canny} is computed from Sobel gradient magnitudes followed by non-maximum suppression and hysteresis thresholding, so it stands in for the RGB-derived edge and gradient family (Canny, Sobel, Laplacian, gradient magnitude). The student is an RGB ResNet-34 distilled from this frozen teacher with exactly the ED\_Student setting ($\alpha = 0.2$, $T = 3.0$, 120 epochs).

\paragraph{Synth\_Distilled (augmentation-difference event proxy).}
The teacher input is a two-channel ON/OFF map computed from a single still RGB image:
\begin{enumerate}
    \item Draw \texttt{RandomResizedCrop}(224, scale $0.5$--$1.0$) and \texttt{RandomHorizontalFlip} once and share the result between the two views.
    \item Apply \texttt{RandomAffine}(\texttt{translate}$=\pm 3\%$, i.e.\ $\pm 7$ px at 224\,px) and \texttt{ColorJitter}(brightness $0.15$, contrast $0.15$, saturation $0.10$, hue $0.05$) to each view with independent random draws.
    \item Convert both views to grayscale (ITU-R\,601) and take the difference $d = g_2 - g_1$.
    \item Set $\mathrm{ON} = \max(d, 0)$ and $\mathrm{OFF} = \max(-d, 0)$, and stack them into a two-channel tensor normalized with mean $(0,0)$ and standard deviation $(0.5,0.5)$.
\end{enumerate}
Sharing the crop is deliberate. When the crop is drawn independently for the two views, the difference degenerates into broad double-exposure blobs that do not follow object contours; sharing it leaves the $\pm 7$ px translation and the mild photometric jitter as the only sources of difference, which yields sharp ON/OFF responses along contours. The teacher is a ResNet-34 whose \texttt{conv1} is replaced by a two-channel input layer (\texttt{out\_channels}$=64$, kernel $7$, stride $2$, padding $3$), trained with the same optimizer, precision, batch size and schedule as above; the maps are generated on the fly inside the data loader. The student again follows the ED\_Student setting. We stress that this proxy is derived from still RGB images: it is neither the output of an event simulator nor of a real event camera.

\begin{table}[h]
\centering
\small
\begin{tabular}{llcc}
\toprule
Control & Teacher input & Teacher & Student \\
\midrule
Canny\_Distilled & Canny edge map (3ch) & 55.82 & 71.15 \\
Synth\_Distilled & augmentation-difference ON/OFF (2ch) & 64.52 & 71.78 \\
\bottomrule
\end{tabular}
\caption{Clean top-1 accuracy (\%) of the two additional non-event controls. Teacher accuracy is measured on the correspondingly transformed ImageNet validation split; student accuracy is on the standard RGB validation split.}
\label{tab:added-controls}
\end{table}

\subsection{SOTA Training Recipe}
\label{sec:ap_sota_recipe}

To check that the reported gains are not an artifact of a weakly trained baseline (Section~\ref{sec:6-4_vs_robust_methods}), we retrained both the baseline and ED\_Student with a timm-style recipe: timm \texttt{resnet34} with \texttt{drop\_path\_rate}$=0.05$ (stochastic depth), label smoothing $0.1$, MixUp $\alpha = 0.2$ applied to the cross-entropy branch only, RandAugment (\texttt{num\_ops}$=2$, \texttt{magnitude}$=9$), RandomErasing $p = 0.25$, 300 epochs, AdamW with weight decay $0.05$, 5 warmup epochs followed by cosine decay, effective batch size 1024, and bf16-mixed precision. The distillation setting of the SOTA ED\_Student is unchanged from ED\_Student ($\alpha = 0.2$, $T = 3.0$).

The learning rate is set to $75\times 10^{-5}$ for both models. This departs from the usual timm default, and we adopt it so that the SOTA baseline and the SOTA ED\_Student differ only in the presence of the distillation term, which is what the comparison is meant to isolate.

\begin{table}[h]
\centering
\small
\begin{tabular}{lcc}
\toprule
& SOTA baseline & SOTA ED\_Student \\
\midrule
Clean top-1 & 75.92 & 70.80 \\
ImageNet-Sketch top-1 & 26.98 & 28.94 \\
Hue rotation, mean over $36$--$324^\circ$ & 63.96 & 65.76 \\
Patch shuffle, grid $=4$ & 53.82 & 37.48 \\
Patch rotation, grid $=4$ & 46.54 & 31.61 \\
FGSM, mean over $\epsilon \in \{0.25, 0.5, 1\}/255$ & 34.74 & 28.70 \\
PGD-10, mean over the same $\epsilon$ & 16.65 & 16.30 \\
Bandpass accuracy drop at $0.95\pi$ (pp, lower is better) & 25.27 & 21.07 \\
Linear probe, mean over 19 tasks & 74.19 & 75.45 \\
\bottomrule
\end{tabular}
\caption{Baseline and ED\_Student retrained under the SOTA recipe. ED's advantage persists on shape, color, high-frequency-band robustness and downstream transfer, but not on the adversarial axes, where the stronger augmentation already makes the SOTA baseline robust at these perturbation sizes.}
\label{tab:sota-recipe}
\end{table}

\subsection{Cross-Method Comparison on the Figure-1 Axes}
\label{sec:ap_central_comparison}

Section~\ref{sec:6-4_vs_robust_methods} uses one metric per axis for compactness. Table~\ref{tab:central_comparison} expands that comparison onto the axes summarized in the right panel of Figure~\ref{fig:method_overview}, reporting two color axes (hue rotation and grayscale) and two shape axes (cue-conflict and ImageNet-Sketch), because each pair probes the same property from a different direction. Synth\_Distilled is compared on the four axes of Table~\ref{tab:joint_profile}.

\begin{table}[h]
\centering
\footnotesize
\setlength{\tabcolsep}{3pt}
\begin{tabular}{lccccccc}
\toprule
& LP mean & Hue Avg. & Grayscale & Cue-conflict & Sketch & Bandpass @$0.95\pi$ & Adv.\ AUC \\
\midrule
baseline & 73.5 & 61.8 & 62.4 & 0.212 & 23.7 & 44.3 & 6.2 \\
ED\_Student & \textbf{75.1} & 65.4 & 64.6 & 0.375 & 28.1 & 57.5 & 8.9 \\
ED\_SoftLabelOnly & --- & 52.4 & 52.6 & 0.487 & 24.5 & 45.8 & 9.1 \\
Canny\_Distilled & 73.7 & \textbf{66.1} & \textbf{65.1} & 0.270 & 27.9 & 35.6 & 7.5 \\
Adversarial Training & 71.7 & 50.6 & 49.5 & \textbf{0.572} & 21.3 & \textbf{64.5} & \textbf{24.1} \\
Noise Training & 72.7 & 62.6 & 63.3 & 0.258 & 25.4 & 58.2 & 11.5 \\
SIN & 72.6 & 62.3 & 62.5 & 0.428 & \textbf{30.8} & 61.8 & 6.5 \\
\bottomrule
\end{tabular}
\caption{Comparison on the axes of Figure~\ref{fig:method_overview}. LP mean: linear-probe accuracy averaged over the 19 tasks of Table~\ref{tab:downstream_task}; the linear probe is run for the five models compared in that table plus Canny\_Distilled. Hue Avg.: top-1 averaged over ten hue-rotation angles including $0^\circ$. Grayscale: top-1 on the grayscale validation split. Cue-conflict: shape bias under the 16-class restricted-argmax protocol. Sketch: ImageNet-Sketch top-1. Bandpass: top-1 under band-limited noise centered at $0.95\pi$ (higher is better). Adv.\ AUC: area under the PGD accuracy curve over $\epsilon \in \{0, 0.25, 0.5, 1\}/255$, scaled by 100. Bold marks the best value in each column.}
\label{tab:central_comparison}
\end{table}

No single supervision dominates. Adversarial Training is the most robust on both the frequency and the adversarial axes but loses 11.2\,pp on hue rotation, 12.9\,pp on grayscale and 1.8\,pp on downstream transfer relative to the baseline. SIN attains the highest top-1 accuracy on ImageNet-Sketch but does not improve downstream transfer. Canny\_Distilled matches ED\_Student on the color and sketch axes, yet its cue-conflict shape bias (0.270) stays much closer to the baseline (0.212) than to ED\_Student (0.375), and it is the only model whose bandpass accuracy falls below the baseline. ED\_Student is the only supervision that improves over the baseline on color, shape, high-frequency-band robustness, and downstream transfer simultaneously.

\section{Extended Analysis of Color Invariance}
\label{sec:ap_color}

In this appendix, we present additional quantitative and qualitative analyses of the Color and Hue Invariance discussed in Section~\ref{sec:4-1_color} of the main paper, which did not fit on the main pages. Specifically, we provide three stages of validation: (i) a comparison of hue-rotation robustness across the nine models of Figure~\ref{fig:ap_hue_all_models} (Appendix~\ref{sec:ap_color_hue}), (ii) behavior under grayscale conditions in which only color information is removed (Appendix~\ref{sec:ap_color_grayscale}), and (iii) an internal-representation analysis via visualization of early-layer feature maps under hue rotation (Appendix~\ref{sec:ap_color_featmap}). Behavior under B\&W, Canny, and silhouette conditions, in which texture is further removed beyond grayscale and only shape information remains, is consolidated in Appendix~\ref{sec:ap_shape}, organized from the perspective of shape bias.

\subsection{Across-Model Comparison under Hue Rotation}
\label{sec:ap_color_hue}

\begin{figure*}
\vspace{-1.5em}
\centering
\includegraphics[width=\linewidth]{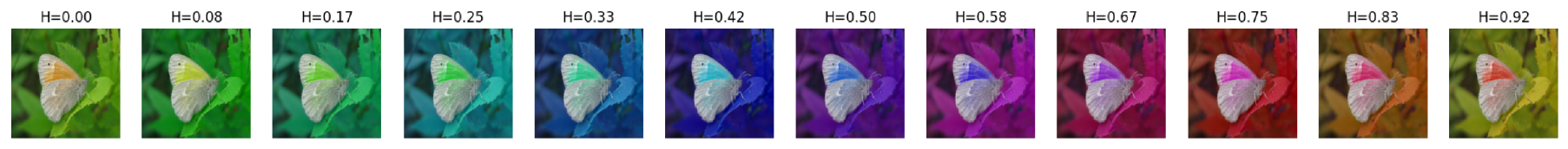}
\vspace{-0.5em}
\caption{Visual examples of hue-rotated inputs. A single ImageNet validation image is shown after rotating its hue from $0^\circ$ to $330^\circ$ in $30^\circ$ steps (12 images). The label $H$ denotes the relative hue value in HSV space ($H \in [0, 1)$, with $1$ corresponding to $360^\circ$).}
\label{fig:ap_hue_rotation_example}
\vspace{0.5em}
\end{figure*}

\begin{figure*}
\vspace{-1.0em}
\centering
\includegraphics[width=\linewidth]{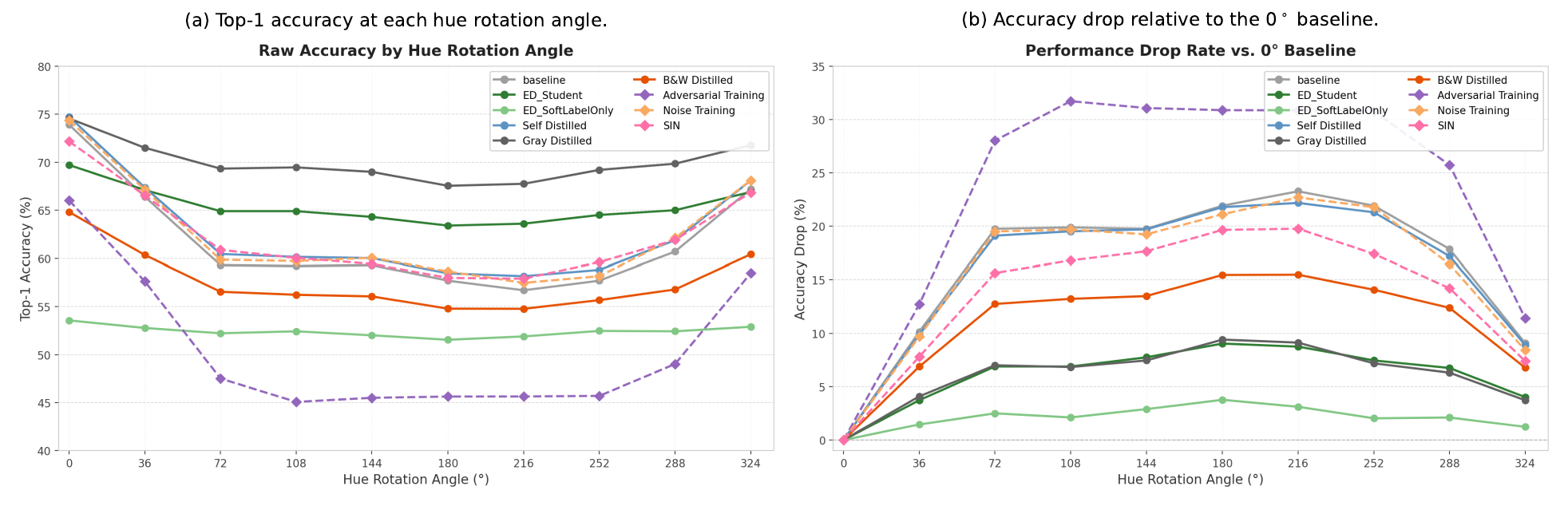}
\vspace{-1.1em}
\caption{Hue-rotation robustness across all 9 models. a) Top-1 accuracy at each hue angle ($0^\circ$--$324^\circ$ in $36^\circ$ steps). b) Accuracy drop relative to $0^\circ$.}
\label{fig:ap_hue_all_models}
\vspace{1em}
\end{figure*}

Figure~\ref{fig:ap_hue_all_models} shows, for all 9 models trained in this work, (a) raw top-1 accuracy at each hue angle and (b) the accuracy drop relative to the $0^\circ$ baseline, when the hue of the entire input image is rotated from $0^\circ$ to $324^\circ$ in $36^\circ$ steps. Figure~\ref{fig:three_phenomenon}~(a) of the main paper shows only 3 of the models (baseline, ED\_Student, ED\_SoftLabelOnly), so here we summarize the behavior of the remaining 6 models (Self Distilled, Gray Distilled, B\&W Distilled, Adversarial Training, Noise Training, SIN).

Adversarial Training shows the largest accuracy drop among all models. The perturbations PGD training assumes are small per-pixel high-frequency noises in the $L_\infty$ norm, whereas hue rotation is a structured global transformation across the entire color space and is therefore outside the scope of the attack model. Moreover, by suppressing reliance on high-frequency texture cues, adversarial training can, as a side effect, strengthen reliance on the remaining color-related cues.

Self Distilled, Noise Training, and the SIN model all show only roughly the same hue-rotation robustness as the baseline. Notably, the SIN model shows little color invariance despite acquiring a strong shape bias via Stylized-ImageNet, indicating that acquiring shape bias does not necessarily entail color invariance.

B\&W Distilled and Gray Distilled, in which color information has been heavily removed on the teacher side in advance, show stronger color invariance than the baseline, but their magnitude remains comparable to or less than that of ED\_Student. This supports that the color invariance acquired by the Event-distilled model cannot be fully explained by inheritance of an inductive bias from removing color information from the teacher alone, and is rooted in dependence on brightness gradients (edges) intrinsic to the event domain (see Section~\ref{sec:6-1_ablation} of the main paper).

\subsection{Robustness under Grayscale Inputs}
\label{sec:ap_color_grayscale}

\begin{table}[!htp]\centering\small
\begin{tabular}{lrrrrrr}\toprule
&ImageNet-val &grayscale &sketch &B\&W &canny &silhouette \\\midrule
baseline &73.90 &62.45 &23.70 &41.39 &10.83 &3.30 \\
ED\_Student &69.70 &\uline{64.64} &\uline{28.08} &\uline{49.01} &\uline{13.27} &\textbf{5.43} \\
ED\_SoftLabelOnly &53.50 &52.58 &24.47 &34.87 &11.45 &\uline{5.42} \\
Self Distilled &\textbf{74.72} &62.97 &24.43 &42.69 &11.59 &3.24 \\
Gray Distilled &\uline{74.60} &\textbf{69.51} &25.60 &45.56 &7.87 &3.63 \\
B\&W Distilled &64.81 &56.20 &23.47 &45.54 &12.12 &3.22 \\
Adversarial Training &66.02 &49.50 &21.27 &29.78 &4.43 &3.28 \\
Noise Training &74.35 &63.32 &25.39 &45.39 &10.28 &4.09 \\
SIN &72.18 &62.53 &\textbf{30.84} &\textbf{51.62} &\textbf{25.77} &5.38 \\
\bottomrule
\end{tabular}
\caption{Top-1 accuracy (\%) on the ImageNet validation set with color/texture information progressively stripped. Bold marks the highest in each column, underline the second. Referenced from both Appendix~\ref{sec:ap_color_grayscale} and \ref{sec:ap_shape_strip}. Visual examples in Figure~\ref{fig:ap_image_sample}.}
\label{tab:quantitative_colorless}
\end{table}

\begin{figure*}
\vspace{-1.5em}
\centering
\includegraphics[width=\linewidth]{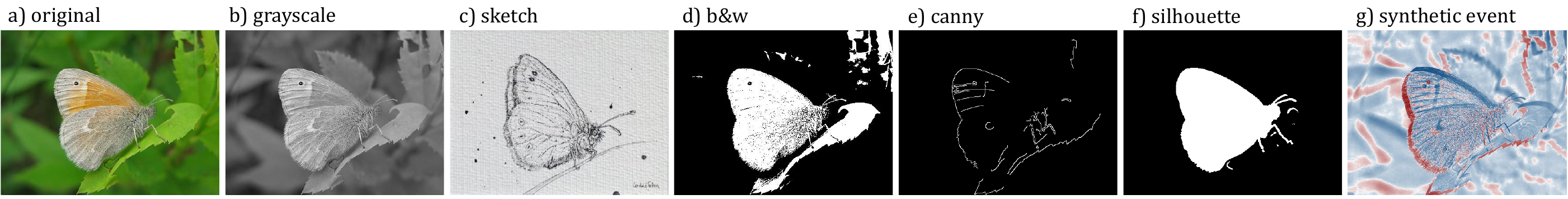}
\vspace{-1.1em}
\caption{Visual examples of each transformation. a) RGB, b) grayscale, c) sketch, d) B\&W (Otsu thresholding), e) Canny, f) silhouette, g) synthetic event. The last panel is the two-channel ON/OFF map used to train the Synth\_Distilled teacher (Section~\ref{sec:ap_added_controls}), rendered with ON in red and OFF in blue; it is the grayscale difference between two augmented views of a) and is not the output of an event simulator or of a real event camera.}
\label{fig:ap_image_sample}
\vspace{1em}
\end{figure*}

Table~\ref{tab:quantitative_colorless} shows top-1 accuracy for all 9 models under conditions in which the ImageNet validation set is preprocessed with five transformations (grayscale, sketch, B\&W, Canny, silhouette). Note that, unlike the others, the sketch column is not a transformation derived from ImageNet-1k but the accuracy on the validation split of the independent ImageNet-Sketch dataset~\cite{wang2019sketch}. In this subsection, we focus on the grayscale column, which removes only color information (the behavior under shape-only conditions following binarization is treated in Appendix~\ref{sec:ap_shape_strip}). Visual examples of each transformation are shown in Figure~\ref{fig:ap_image_sample}.

Under grayscale, Gray Distilled --- in which color information has been removed on the teacher side in advance --- attains the highest accuracy, followed by ED\_Student. Compared to baseline, the Event-distilled model shows a clear advantage, evidencing an inference path that does not require color information. The SIN model is at the baseline level, confirming that acquiring shape bias by another route does not in itself confer color invariance. By contrast, Adversarial Training degrades to roughly 13\,pp below baseline already at the grayscale stage, indicating that the feature-representation distortion accompanying adversarial training becomes apparent under the loss of color information.

\subsection{Layer-wise Feature Maps under Hue Rotation}
\label{sec:ap_color_featmap}

\begin{figure*}
\vspace{-1.5em}
\centering
\includegraphics[width=0.85\linewidth]{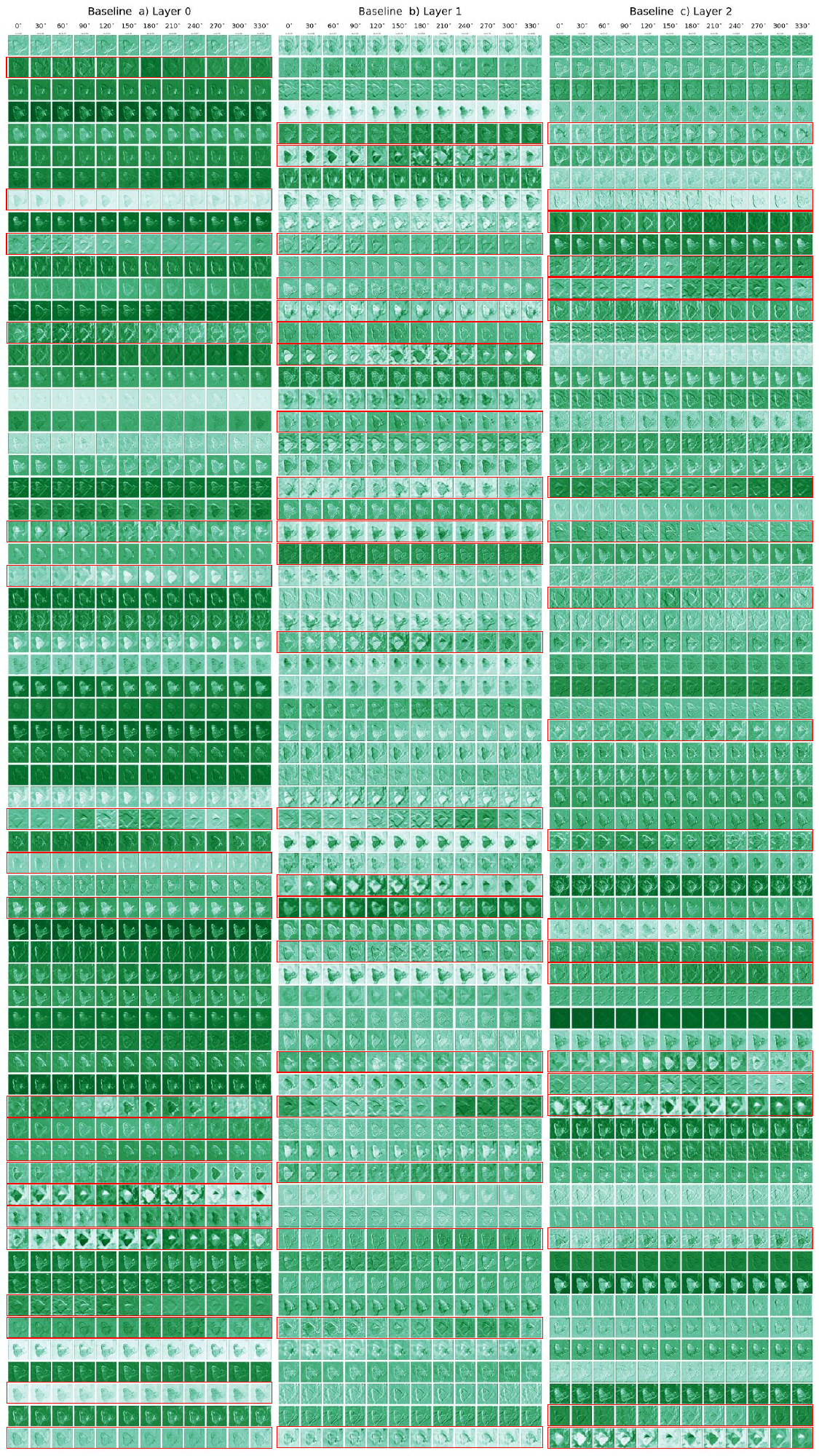}
\caption{Feature maps in the early layers (Layer 0--2) of the baseline model. Each column corresponds to a hue rotation angle ($0^\circ$--$330^\circ$ in $30^\circ$ steps), and each row corresponds to one channel. Red boxes mark channels whose response varies substantially with hue change (not color-invariant). See Figure~\ref{fig:ap_hue_rotation_example} for visual examples of the hue-rotated inputs.}
\label{fig:ap_fig_hue_rotate_map_baseline}
\vspace{-1em}
\end{figure*}

\begin{figure*}
\vspace{-1.5em}
\centering
\includegraphics[width=0.85\linewidth]{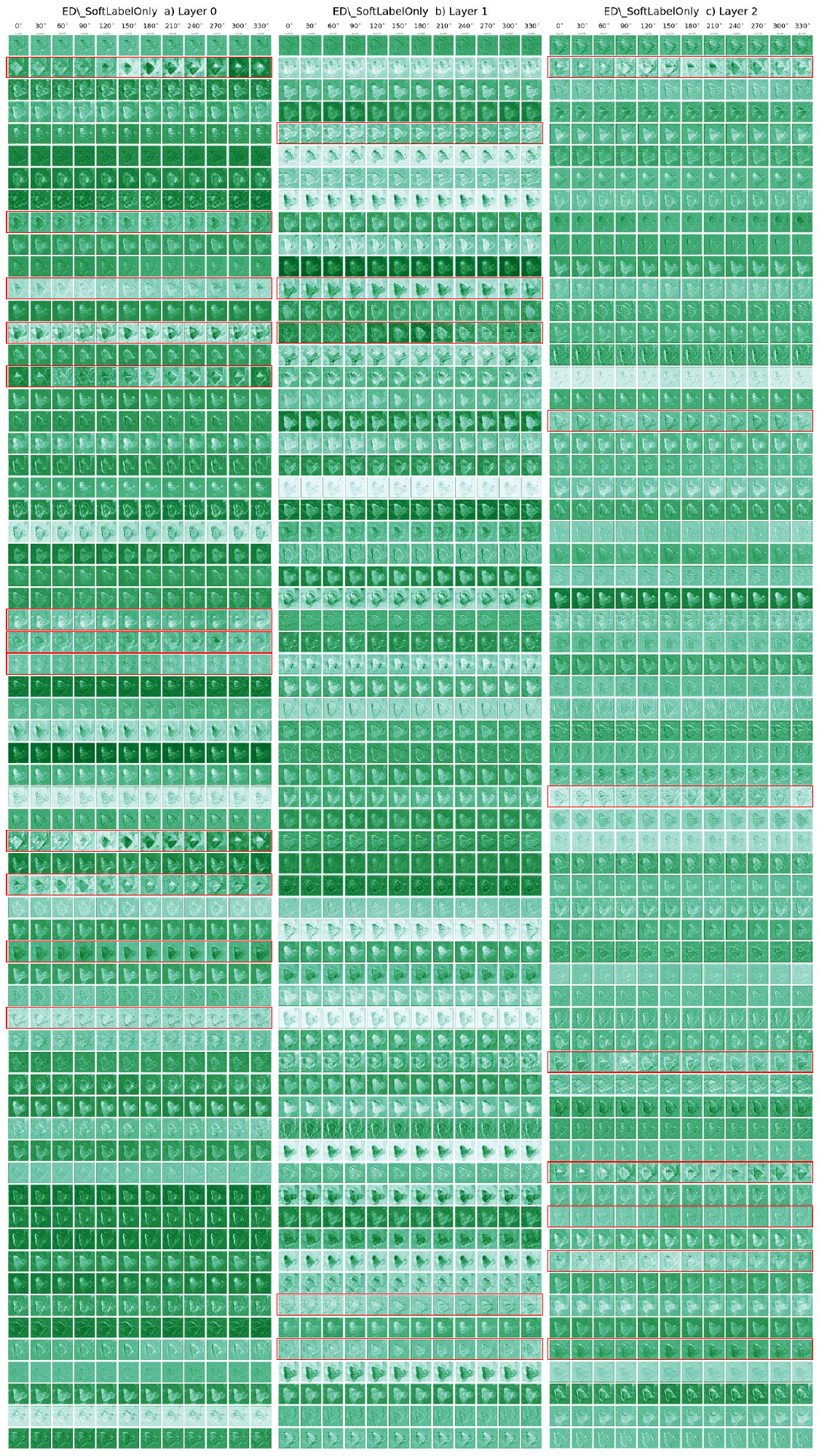}
\caption{Early-layer feature maps of ED\_SoftLabelOnly. Same format as Figure~\ref{fig:ap_fig_hue_rotate_map_baseline}. Responses are notably invariant to hue rotation; only a few channels are flagged in red.}
\label{fig:ap_fig_hue_rotate_map_ED}
\vspace{-1em}
\end{figure*}

To directly confirm at which stage of the representation space the color invariance emerges, we visualized the feature maps in the early convolutional layers (Layer 0--2) of ResNet-34 under hue-rotated inputs. Figure~\ref{fig:ap_fig_hue_rotate_map_baseline} shows the baseline model and Figure~\ref{fig:ap_fig_hue_rotate_map_ED} the ED\_SoftLabelOnly model: each figure displays the responses of every layer and channel as the hue of the input image is rotated from $0^\circ$ to $330^\circ$ in $30^\circ$ steps. The red boxes in each figure highlight channels whose response strength or spatial distribution varies substantially with the input hue --- that is, channels without color invariance.

In the baseline model, many channels already at Layer~0 are flagged in red, indicating responses strongly dependent on input color information. In contrast, in the ED\_SoftLabelOnly model, the early-layer feature maps remain almost invariant under the same hue-rotated inputs, and only a very small number of channels are flagged in red. In other words, this model is unaffected by changes in input color and consistently extracts only geometric structure (edges) based on the Structural Prior discussed in Section~\ref{sec:5-1} of the main paper. This is confirmed at the layer level via visualization. The result demonstrates, at the per-pixel response level, that the color invariance of the Event-distilled model arises not from post-processing at the output layer but from a representation transformation at the stage closest to the input.

\section{Extended Analysis of Shape Bias}
\label{sec:ap_shape}

In this appendix, we provide additional quantitative and qualitative analyses of the High Shape Bias discussed in Section~\ref{sec:4-2_shape} of the main paper, which did not fit on the main pages. Specifically, we examine the topic from four perspectives: (i) cross-architecture and cross-model comparison of cue-conflict experiments (Appendix~\ref{sec:ap_shape_cueconflict}), (ii) accuracy comparison under shape-only conditions in which color and texture information have been stripped to the extreme (Appendix~\ref{sec:ap_shape_strip}), (iii) behavior when the macroscopic shape is destroyed by patch-shuffle / patch-rotation (Appendix~\ref{sec:ap_shape_patchshuffle}), and (iv) internal-representation analysis through a functional taxonomy of first-layer convolutional filters (Appendix~\ref{sec:ap_shape_filter}).

\subsection{Cue-Conflict: Across Architectures and Models}
\label{sec:ap_shape_cueconflict}

\begin{figure*}
\vspace{-1.5em}
\centering
\includegraphics[width=\linewidth]{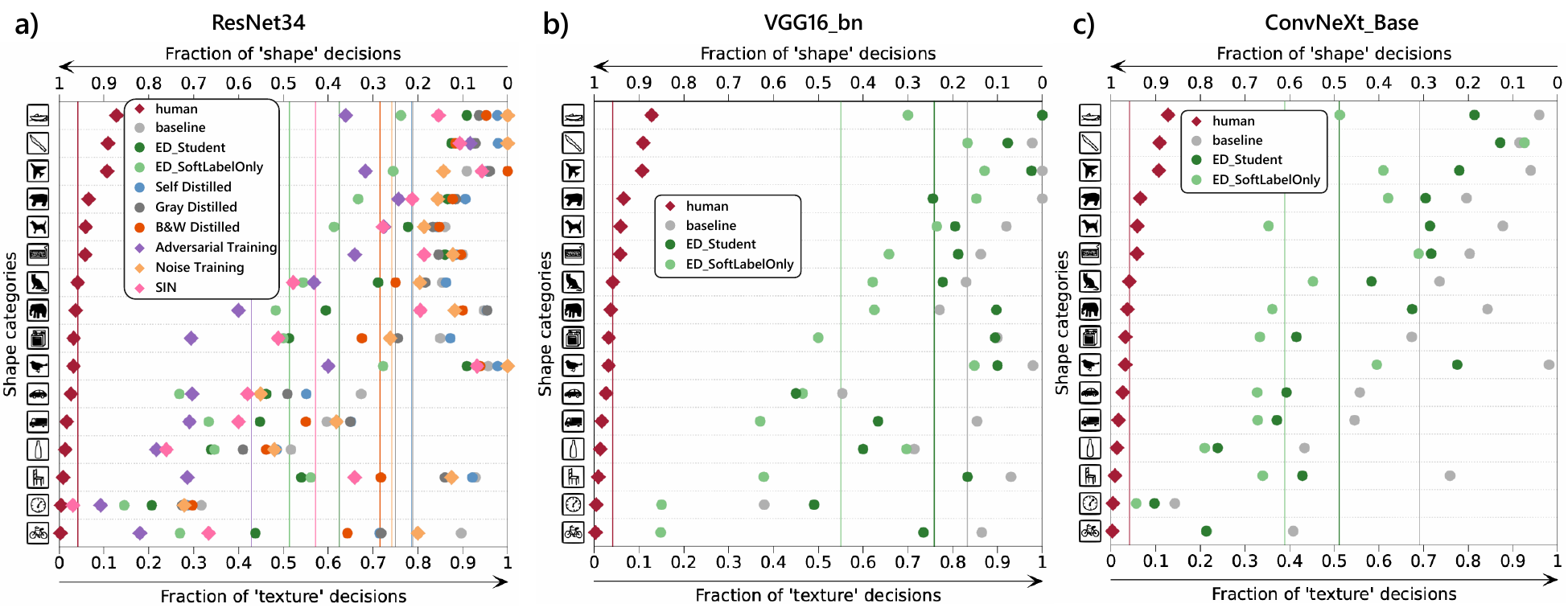}
\vspace{-1.1em}
\caption{Cue-conflict experiments across architectures and models. a) Shape-bias matrixplot of all 9 ResNet-34 models (baseline, ED\_Student, ED\_SoftLabelOnly, Self/Gray/B\&W Distilled, Adversarial Training, Noise Training, SIN) and the human reference, computed using the bethgelab/model-vs-human 16-class restricted-argmax protocol~\cite{geirhos2019texture_bias}. b) VGG-16 and c) ConvNeXt-Base, each with the three models baseline / ED\_Student / ED\_SoftLabelOnly. Each panel plots the texture-versus-shape decision ratio per category as a scatter plot.}
\label{fig:ap_cue_conflict_archs}
\vspace{-1em}
\end{figure*}

We extend the cue-conflict result on ResNet-34 shown in Figure~\ref{fig:three_phenomenon}~(b) of the main paper along two axes: (i) across architectures (VGG-16, ConvNeXt-Base) and (ii) across models within ResNet-34 (all 9 models).

\textbf{Across architectures (Figure~\ref{fig:ap_cue_conflict_archs}b, c).}
For both VGG-16 and ConvNeXt-Base, the Event-distilled models (ED\_Student, ED\_SoftLabelOnly) shift clearly toward the shape side (left) of the plot relative to baseline, matching the trend on ResNet-34 shown in Figure~\ref{fig:three_phenomenon}~(b) of the main paper. This result constitutes one piece of evidence for the architectural generality discussed in Section~\ref{sec:6-2_architectural_generality} of the main paper.

\textbf{Across models on ResNet-34 (Figure~\ref{fig:ap_cue_conflict_archs}a).}
Under the official 16-class restricted-argmax protocol, the overall shape bias is ordered as Adversarial Training (0.572) $>$ ED\_SoftLabelOnly (0.487) $>$ SIN (0.428) $>$ ED\_Student (0.375) $>$ B\&W Distilled (0.284) $>$ Noise Training (0.258) $>$ Gray Distilled (0.251) $>$ Self Distilled (0.214) $>$ baseline (0.212), with the human reference at 0.959. SIN sits between ED\_Student and ED\_SoftLabelOnly, so the strength of its shape bias is comparable to that of the Event-distilled models. However, the Event-distilled model differs from both Adversarial Training and SIN in its side-effect profile: Adversarial Training exhibits the largest accuracy drop under hue rotation among all models (Appendix~\ref{sec:ap_color_hue}) and falls below the baseline on 12 of 19 downstream transfer tasks (Table~\ref{tab:downstream_task}); SIN's hue-rotation robustness is at the baseline level (Appendix~\ref{sec:ap_color_hue}) and 12 of 19 downstream tasks fall below baseline, while it achieves the highest accuracy on the shape-only inputs (B\&W / Canny / sketch columns of Table~\ref{tab:quantitative_colorless}), even surpassing ED\_Student. Unlike these two routes, the Event-distilled model preserves color invariance (Appendix~\ref{sec:ap_color}) and transfer performance (Table~\ref{tab:downstream_task}) while still acquiring a strong shape bias, and therefore offers an advantage from a holistic perspective that includes not just the magnitude of the shape bias but also a more favorable trade-off across the evaluated metrics.

\subsection{Robustness under Shape-Only Inputs}
\label{sec:ap_shape_strip}

Of Table~\ref{tab:quantitative_colorless} and Figure~\ref{fig:ap_image_sample} shown in Appendix~\ref{sec:ap_color_grayscale}, this subsection focuses in particular on the four conditions in which most texture and color information have been stripped and only shape information remains (sketch, B\&W, Canny, silhouette).

Under these four conditions, the SIN model --- whose shape learning is directly induced by Stylized-ImageNet co-training --- attains the highest accuracy on three of them (sketch, B\&W, Canny), with ED\_Student following as second on each. On silhouette, ED\_Student is highest, followed essentially in a tie by ED\_SoftLabelOnly and SIN. Even though the Event-distilled model performs no explicit shape learning, its ability to infer class from shape information alone reaches a level comparable to SIN, and clearly exceeds Self Distilled and Noise Training, both of which use RGB training inputs under all conditions. This corroborates --- including the remaining control models that did not fit in the main paper --- the claim of Section~\ref{sec:5-3} of the main paper that the Event-distilled model can perform stable class inference from object shape information alone, without requiring superficial cues such as color or texture.

In contrast, Adversarial Training falls below baseline on all four shape-extraction conditions. The fact that adversarial training, despite having the effect of suppressing high-frequency texture dependence, does not lead to inference from shape information suggests that the essence of robustness lies not only in texture suppression but in selective acquisition of shape features.

\subsubsection{Generation of Canny and Silhouette Evaluation Inputs}
\label{sec:ap_canny_silhouette}

The Canny and silhouette columns of Table~\ref{tab:quantitative_colorless} are evaluated on offline-generated parallel trees of the ImageNet validation split (one image per source image, saved as PNG to avoid JPEG-compression artifacts). The B\&W column is generated by the Otsu thresholding pipeline already documented in Section~\ref{sec:aug-bw-pipe}.

\textbf{Canny.} Each ImageNet validation image is converted to grayscale (\texttt{PIL.Image.convert(``L'')}) and normalized to float32 in $[0, 1]$. A Gaussian blur with $\sigma = 1.0$ and kernel size $\lceil 6\sigma + 1 \rceil = 7$ (forced odd) is applied, after which the image is rescaled to \texttt{uint8} and passed to \texttt{cv2.Canny} with low/high thresholds $0.35\!\times\!255 \approx 89$ and $0.9\!\times\!255 \approx 230$. The resulting binary edge map is saved as PNG.

\textbf{Silhouette.} Each image is loaded as RGB and passed through the \texttt{rembg} library with the \texttt{u2net} session for foreground extraction. The returned RGBA tensor's alpha channel is binarized by the rule $\mathrm{alpha} > 10 \mapsto 255$, $\mathrm{alpha} \le 10 \mapsto 0$, yielding a two-valued mask that is saved as PNG.

Both pipelines preserve the strict 0/255 binary structure of the saved images, allowing the validation loader to feed them directly into the same evaluation transform (\texttt{Resize}(256) + \texttt{CenterCrop}(224) + ImageNet \texttt{Normalize}) used for the RGB / grayscale columns.

\subsection{Patch-Shuffle and Patch-Rotation Experiments}
\label{sec:ap_shape_patchshuffle}

\begin{figure*}
\vspace{-1.5em}
\centering
\includegraphics[width=\linewidth]{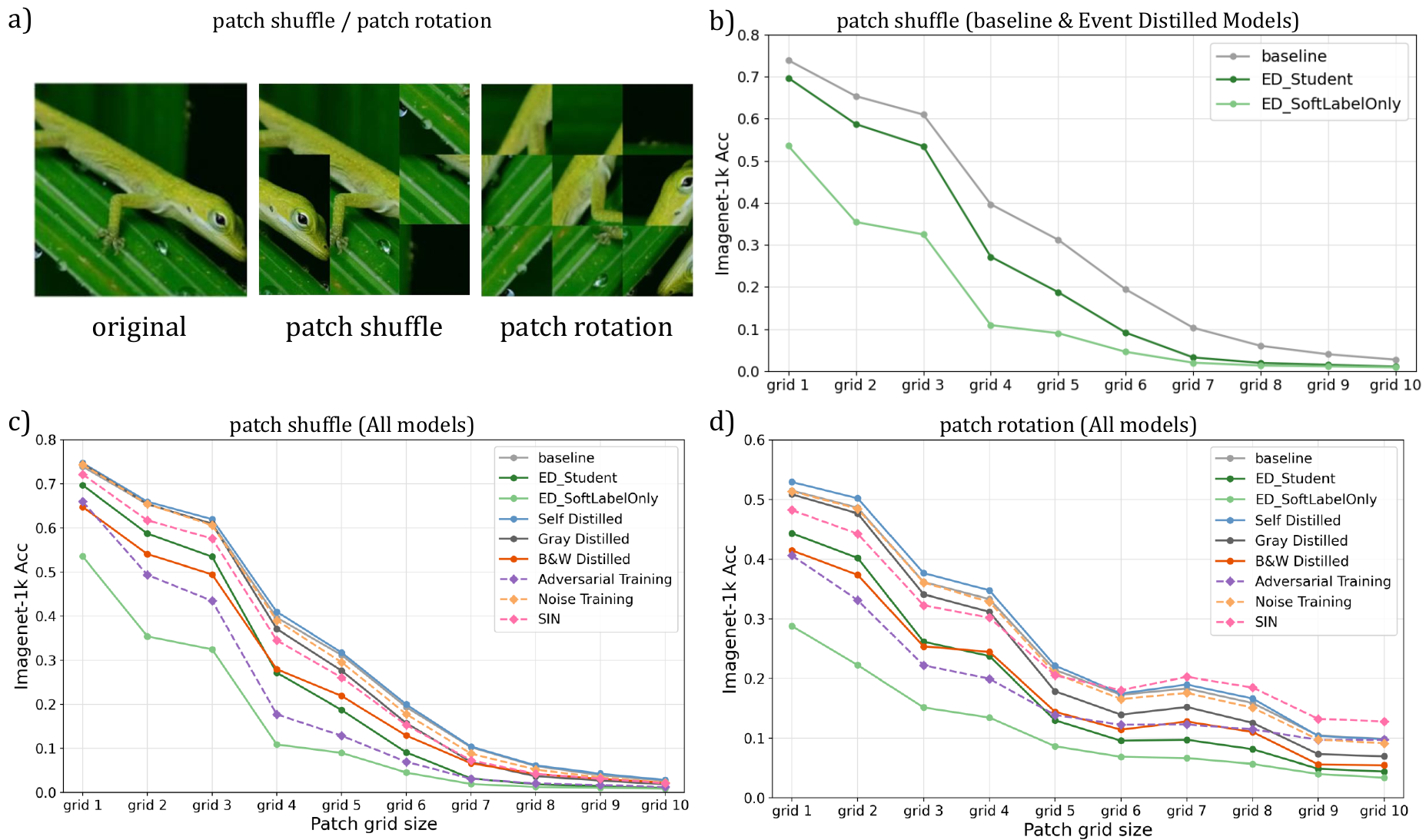}
\vspace{-1.1em}
\caption{Patch-shuffle / patch-rotation experiments. a) Visual examples. b) Top-1 accuracy under patch shuffle as a function of grid granularity (baseline vs.\ Event-distilled models). c) Accuracy under patch shuffle for all 9 models. d) Same for patch rotation.}
\label{fig:ap_patch_shuffle}
\vspace{1em}
\end{figure*}

As further evidence of shape bias, we conducted experiments that destroy the continuity of macroscopic edges by dividing the image into a grid and shuffling / rotating each patch. Patch shuffling and rotation are widely adopted diagnostic procedures: they preserve the local texture inside each patch while selectively destroying the global shape information that spans across patches, allowing one to disentangle whether a model relies on within-patch local cues or on across-patch macroscopic shape~\cite{naseer2021intriguing}. By varying the grid granularity, we quantitatively evaluate the spatial scale of the cues on which the Event-distilled model relies. Figure~\ref{fig:ap_patch_shuffle} shows the top-1 accuracy progression as the patch size (grid granularity) is varied.

When the grid is coarse, the local texture inside each patch is preserved, so the texture-dependent baseline maintains relatively high accuracy. However, as the grid becomes finer and the continuity of macroscopic edges is broken, the Event-distilled model's accuracy drops more sharply than the baseline's. This is direct evidence that the information source on which the Event-distilled model relies for inference is not local texture within a patch but the macroscopic shape structure that holds across patches. The same trend is observed under patch rotation, and in both conditions the Event-distilled model exhibits steeper accuracy degradation than even Gray Distilled and B\&W Distilled, which are distilled from teachers stripped of color information. This confirms that the Event-distillation-specific dependence on macroscopic shape continuity is a property distinct from the other distillation variants. The SIN model, which acquires shape bias through Stylized-ImageNet, shows accuracy curves comparable to the baseline under both patch-shuffle and patch-rotation conditions, indicating that even when shape bias is acquired through an alternative route, sensitive dependence on macroscopic shape continuity does not necessarily follow.

\begin{figure*}
\vspace{-1.5em}
\centering
\includegraphics[width=\linewidth]{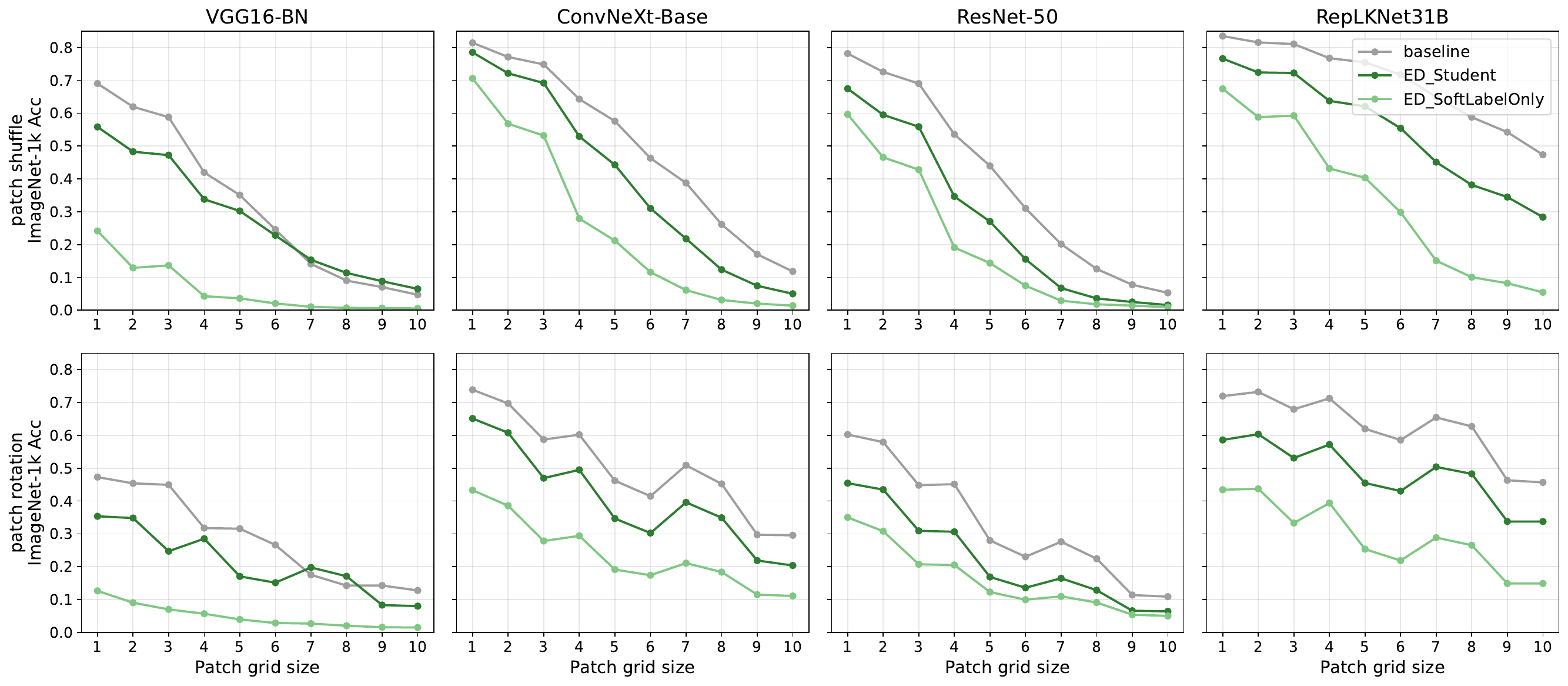}
\vspace{-1.1em}
\caption{Cross-architecture comparison of patch-shuffle / patch-rotation experiments. For VGG-16, ConvNeXt-Base, ResNet-50, and RepLKNet31B, we report the top-1 accuracy progression of baseline / ED\_Student / ED\_SoftLabelOnly as the number of patch divisions varies. Top: patch shuffle. Bottom: patch rotation.}
\label{fig:ap_patch_archs}
\vspace{1em}
\end{figure*}

As part of the architectural generality validation discussed in Section~\ref{sec:6-2_architectural_generality} of the main paper, we also conduct the same patch-shuffle / patch-rotation experiments on VGG-16, ConvNeXt-Base, ResNet-50, and RepLKNet31B. As shown in Figure~\ref{fig:ap_patch_archs}, across all architectures the Event-distilled model exhibits the same characteristic behavior as on ResNet-34: its accuracy drops more sharply than baseline as the patch size becomes finer. This confirms that the Event-distilled model's dependence on macroscopic shape continuity arises consistently regardless of the architectural design.

\subsection{Functional Taxonomy of First-Layer Filters}
\label{sec:ap_shape_filter}

\begin{figure*}
\vspace{-1.5em}
\centering
\includegraphics[width=\linewidth]{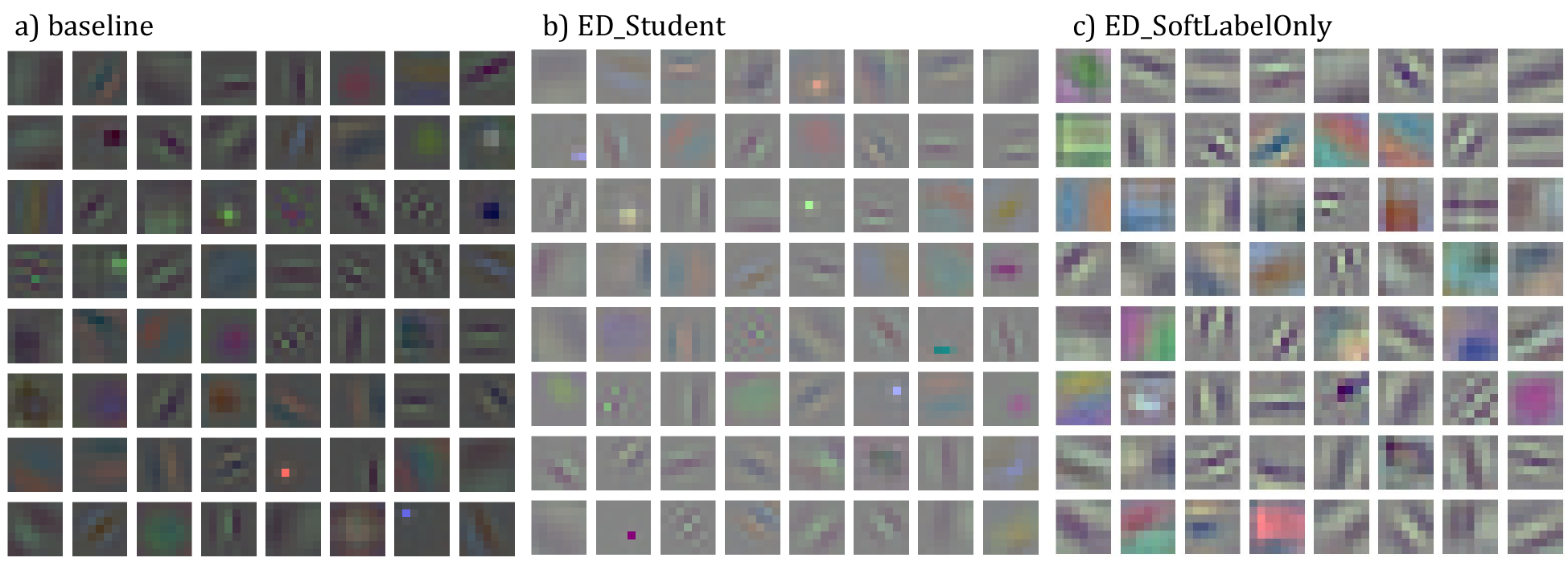}
\vspace{-1.1em}
\caption{All 64 channels of the first convolutional layer of ResNet-34. a) baseline, b) ED\_Student, c) ED\_SoftLabelOnly. Whereas Figure~\ref{fig:filter_jpeg}~(a) shows excerpted channels, this figure shows the layer-wide tendency.}
\label{fig:ap_fig_cnn_filter}
\vspace{-1em}
\end{figure*}

In this subsection, we provide a more detailed analysis of the functional taxonomy of first-layer convolutional filters introduced in Section~\ref{sec:5-1} of the main paper. While Figure~\ref{fig:filter_jpeg}~(a) of the main paper shows only a few excerpted channels due to space constraints, Figure~\ref{fig:ap_fig_cnn_filter} visualizes all 64 channels of the first convolutional layer of each model. The Event-distilled model (especially ED\_SoftLabelOnly) is observed to dominantly acquire smooth, spatially highly correlated structures across the entire filter layer, not just an excerpt.

The acquired 64 filters can be broadly classified into the following three types based on their spatial and color response characteristics.

\begin{itemize}
    \item Monochromatic Global Filters (e.g., 1st column of Figure~\ref{fig:filter_jpeg}~(a)): These form smooth, monochromatic gradients that span the entire receptive field, and serve to smooth out local color variation, thereby estimating broad illumination conditions or macroscopic color distribution of the background. They are believed to contribute to adaptive normalization that highlights structural luminance changes in subsequent layers.
    \item Color-Opponent Structural Filters (e.g., 2nd column of Figure~\ref{fig:filter_jpeg}~(a)): These have a structure smoothly partitioned by two near-complementary colors, exhibiting response characteristics analogous to color-opponent responses in the human visual system. They suppress fine textures while extracting macroscopic boundaries between objects with different color characteristics.
    \item Gabor-like Edge Detectors (e.g., 3rd--4th columns of Figure~\ref{fig:filter_jpeg}~(a)): These have orientations and spatial-frequency characteristics similar to Gabor filters, retaining sensitivity to the steep structural gradients (edges) that constitute shape, while suppressing unstructured high-frequency texture. They faithfully reflect the brightness gradients intrinsic to the event sensor and directly extract macroscopic object contours (edge-based shape feature).
\end{itemize}

For quantitative support, the filters' channel-wise correlation (RGB Correlation) and spatial autocorrelation (Spatial Autocorrelation) are shown in Table~\ref{tab:filter_correlation}. ED\_SoftLabelOnly's filters have substantially higher channel correlation and spatial autocorrelation than baseline ($0.747$ vs.\ $0.549$ for RGB Correlation; $0.575$ vs.\ $0.391$ for Spatial Autocorrelation). ED\_Student's first-layer correlations are close to baseline ($0.544$ / $0.398$); we discuss this dichotomy in Appendix~\ref{sec:ap_filter_aug_confound}, where we argue that the strong CE branch dominates first-layer shaping for ED\_Student while the underlying mechanism is distributed across the representation rather than localized in the first layer. The high correlation of ED\_SoftLabelOnly indicates (i) that all RGB channels synchronously extract the same geometric structure, and (ii) that the model has acquired a smooth spatial structure that emphasizes continuity between adjacent pixels, quantitatively reinforcing the structural prior discussed in the main paper.

\begin{table}[!htp]\centering
\scriptsize
\begin{tabular}{lrrr}\toprule
Correlation Type &RGB Correlation &Spatial Autocorrelation \\\midrule
baseline &0.549 &0.391 \\
ED\_Student &0.544 &0.398 \\
ED\_SoftLabelOnly &0.747 &0.575 \\
\bottomrule
\end{tabular}
\caption{Correlation values for the 64 channels of the first convolutional layer of ResNet-34. RGB Correlation is the cross-channel correlation, and Spatial Autocorrelation is between adjacent pixels. Higher values indicate smoother, more isotropic filter structures.}
\label{tab:filter_correlation}
\end{table}

\subsubsection{Robustness of the Mechanistic Claim to the Augmentation Pipeline}
\label{sec:ap_filter_aug_confound}

Since ED\_SoftLabelOnly is trained entirely via the KD branch (the weak pipeline of Table~\ref{tab:aug-rgb-weak}), one might wonder whether the absence of RandAugment and RandomErasing alone --- rather than event distillation --- is what produces the smoother filters and higher correlation values reported in Section~\ref{sec:5-1} of the main paper. We argue that this confound is bounded for three reasons.

\textbf{(A) The augmentation pipeline is a perturbation modulator, not a smoothing driver.} A weak pipeline does not actively push filters toward smoothness; it merely omits perturbations. The gradient signal that organizes the filters comes from the loss target --- here, the event teacher's KL divergence --- not from the augmentation itself. RandAugment~\cite{cubuk2020randaugment} and RandomErasing~\cite{zhong2020random_erasing} are designed to broaden the input distribution rather than to bias filters toward high-frequency texture, so their removal is unlikely on its own to account for the smooth, oriented Gabor-like structures observed in Figure~\ref{fig:ap_fig_cnn_filter}.

\textbf{(B) Filter shape is determined by the loss target and the input statistics, not by the augmentation strength.} Classical work in unsupervised learning of natural image statistics shows that Gabor-like filters emerge from edge-rich input distributions under a sparse-coding objective~\cite{olshausen1996emergence}. CNN first-layer visualizations across diverse training recipes --- from AlexNet~\cite{krizhevsky2012imagenet} to deconvolutional analyses~\cite{zeiler2014visualizing} --- consistently show that texture-tuned first-layer filters arise under standard CE training on natural images, regardless of whether strong augmentation is applied. The Gabor-organized, edge-oriented structure of ED\_SoftLabelOnly therefore reflects the upstream input statistics of N-ImageNet (a flatter PSD and edge-richer distribution; Appendix~\ref{sec:ap_psd}) propagated through the KL distillation, not the absence of augmentation. This is consistent with prior work on the origins of texture bias~\cite{geirhos2019texture_bias,hermann2020origins}, which identifies the loss / data interaction --- not augmentation noise --- as the primary driver of low-level representation shape.

\textbf{(C) The mechanistic claim is not localized in first-layer scalar correlation.} ED\_Student receives 80\% strong-pipeline CE and only 20\% KL; its first-layer scalar correlations (RGB 0.544, Spatial 0.398) are accordingly close to baseline (0.549 / 0.391), which is consistent with the strong CE branch dominating first-layer shaping; we did not measure the relative gradient contributions of the two terms. Yet ED\_Student still exhibits the full set of behavioral signatures attributed to the structural prior: color invariance (Section~\ref{sec:4-1_color}), shape bias (Section~\ref{sec:4-2_shape}), high-frequency-noise robustness (Section~\ref{sec:4-3_highfreq}), the JPEG / pixelate trade-off (Section~\ref{sec:5-2}), shape-only inference (Section~\ref{sec:5-3}), and downstream transfer (Section~\ref{sec:6-3_downstream}). The mechanism is therefore a representation shift that is not localized in first-layer smoothness; the filter analysis serves as one diagnostic among several converging pieces of evidence.

\subsubsection{Where the Filter Change Is Localized}
\label{sec:ap_filter_spectra}

Figure~\ref{fig:filter_jpeg}~(a) shows individual first-layer filters, whereas Figure~\ref{fig:filter_jpeg}~(b) aggregates scalar statistics over all of them, and the two do not obviously agree. A layer-by-layer measurement resolves the tension.

\paragraph{Per-layer alignment between students.}
For each of the 36 convolutional layers of the students we compute, for every filter of one model, the maximum cosine similarity to the filters of another model, and average. At \texttt{conv1} all pairs are high (0.53--0.58), reflecting the shared RGB stem. At \texttt{layer1} the pair ED\_Student--ED\_SoftLabelOnly reaches 0.161, above every other pair (baseline--ED\_Student 0.142, baseline--ED\_SoftLabelOnly 0.135, baseline--Canny\_Distilled 0.144, ED\_Student--Canny\_Distilled 0.141, ED\_SoftLabelOnly--Canny\_Distilled 0.136): the two event-supervised students are measurably closer to each other than to either the baseline or the edge-map control. The gap narrows at \texttt{layer2} and is indistinguishable at \texttt{layer3} and \texttt{layer4}.

The structural change induced by event supervision is therefore localized in the early layers, principally \texttt{layer1}. This layer-wise similarity is a separate diagnostic from the first-layer correlation statistics of Figure~\ref{fig:filter_jpeg}~(b), which aggregate over the filters of a single layer, and the two should not be read as measuring the same quantity.

\section{Extended Analysis of High-Frequency Robustness}
\label{sec:ap_high_freq}

In this appendix, we present additional analyses of Robustness to High-Frequency Perturbations discussed in Section~\ref{sec:4-3_highfreq} of the main paper, which did not fit on the main pages. Specifically, we examine four aspects: (i) hyperparameters used for the PGD evaluation (Appendix~\ref{sec:ap_pgd_eval_details}), (ii) response to band-limited random noise (Appendix~\ref{sec:ap_high_freq_fgsm}), (iii) comprehensive evaluation against all 19 ImageNet-C corruptions (Appendix~\ref{sec:ap_high_freq_inc}), and (iv) interpretation of the Event-distilled model's vulnerability based on the frequency characteristics of additive noise (Appendix~\ref{sec:ap_high_freq_noise_decomp}).

\subsection{PGD Evaluation Hyperparameters}
\label{sec:ap_pgd_eval_details}

The PGD adversarial-robustness curves shown in Figure~\ref{fig:three_phenomenon}~(c) of the main paper and in Figure~\ref{fig:across_model_adv} (Section~\ref{sec:6-2_architectural_generality}) use a common evaluation pipeline, distinct from the PGD recipe used to train the Adversarial Training baseline (Appendix~\ref{sec:ap_a33}, $\epsilon=8/255$, $\alpha=2/255$, $K=10$).
\begin{itemize}
    \item \textbf{Attack}: $L_\infty$ PGD with cross-entropy loss and sign-of-gradient updates.
    \item \textbf{Iterations}: $K = 10$ steps.
    \item \textbf{Step size}: $\alpha = (2.5 \cdot \epsilon) / K$, a conventional choice that slightly overshoots so that the projection is hit at every step independently of $\epsilon$.
    \item \textbf{Random restart}: a single restart, with the perturbation initialized uniformly in $[-\epsilon, +\epsilon]$ and clipped back to $[0, 1]$ before iteration $1$.
    \item \textbf{Image space}: the attack operates in the $[0, 1]$ pixel space; ImageNet normalization is performed inside a model wrapper (\texttt{NormalizedModel}) so that the same attack budget $\epsilon$ corresponds to the same physical pixel perturbation across all compared models, irrespective of training-time normalization statistics.
    \item \textbf{Evaluation set}: the standard ImageNet validation split, processed by \texttt{Resize}(256) + \texttt{CenterCrop}(224) + \texttt{ToTensor} (no normalization outside the wrapper).
    \item \textbf{$\epsilon$ grid}: $\{0,\, 0.25/255,\, 0.5/255,\, 1/255\}$ in Figure~\ref{fig:three_phenomenon}~(c) and Figure~\ref{fig:across_model_adv} (the FGSM panels share the same $\epsilon$ grid). Larger budgets are not reported because all models collapse beyond $1/255$ in this single-restart, low-iteration setting.
\end{itemize}
The same hyperparameters are used for every model in the FGSM/PGD comparisons; only the model weights differ across runs.

\subsection{Frequency-Band Random Noise}
\label{sec:ap_high_freq_fgsm}

\begin{figure*}
\vspace{-1.5em}
\centering
\includegraphics[width=\linewidth]{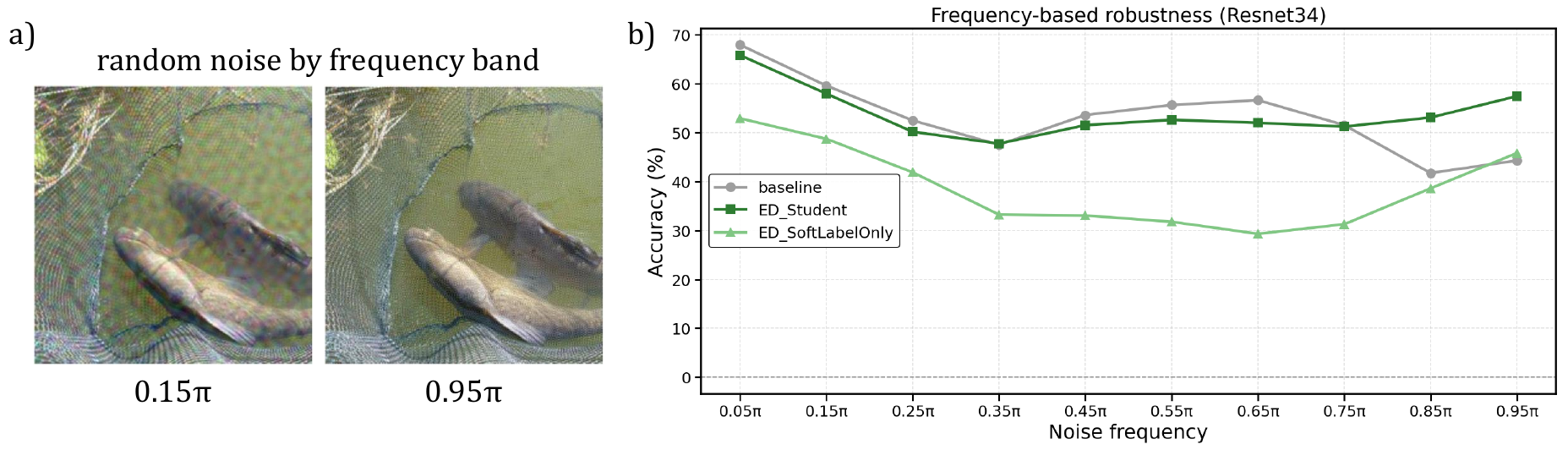}
\vspace{-1.1em}
\caption{Robustness to frequency-band random noise. a) Examples of inputs perturbed by band-limited noise (\emph{left}: center frequency $0.15\pi$, \emph{right}: $0.95\pi$). b) Top-1 accuracy (absolute) under noise concentrated at each band; horizontal axis is the center frequency of the added noise ($0$--$\pi$).}
\label{fig:ap_high_freq_robustness}
\vspace{-1em}
\end{figure*}

Following the procedure of~\citet{park2022vit_freq}, given an input image $x_0$, we generate a perturbed input $x_{\mathrm{noise}}$ that contains only a specified frequency band of noise as
\begin{equation}
x_{\mathrm{noise}} = x_0 + \mathcal{F}^{-1}\!\left( \mathcal{F}(\delta) \odot M_f \right),
\label{eq:freq_noise}
\end{equation}
where $\mathcal{F}(\cdot)$ and $\mathcal{F}^{-1}(\cdot)$ are the 2D Fourier transform and its inverse, $\delta$ is Gaussian noise, $M_f$ is a band-pass mask that lets only the band centered at $f \in [0, \pi]$ pass, and $\odot$ denotes element-wise product. Applying $x_{\mathrm{noise}}$ to the ImageNet validation set and measuring the top-1 accuracy as $f$ varies allows us to evaluate which frequency-band noise each model is fragile or robust to.

The concrete settings are $\epsilon = 0.2$ for the noise magnitude, an $L_\infty$ band-pass mask (a square ring in the frequency plane), and a bandwidth of $0.1\pi$; the eleven center frequencies are $f/\pi \in \{0, 0.05, 0.15, \dots, 0.95\}$. These choices matter: replacing the $L_\infty$ ring with an $L_2$ (circular) one and halving $\epsilon$ shrinks the measured drops by roughly two orders of magnitude, so band-pass numbers are only comparable within a fixed protocol.

Figure~\ref{fig:ap_high_freq_robustness}~(a) shows examples of noisy inputs at low ($0.15\pi$) and high ($0.95\pi$) center frequencies. The low-frequency noise gently modulates the global brightness of the image, whereas the high-frequency noise appears as fine granular patterns on the surface.
Figure~\ref{fig:ap_high_freq_robustness}~(b) plots the absolute top-1 accuracy as the noise's center frequency is varied from $0\pi$ to $\pi$. In particular, in the high-frequency band ($0.75\pi$ and above), the Event-distilled model degrades markedly less than the baseline: at $0.95\pi$ its accuracy drops by 12.2\,pp against the baseline's 29.6\,pp. Both models are still affected, so the difference is one of degree rather than of invariance. This is consistent with the smooth low-pass-filter-like response discussed in Section~\ref{sec:5-1} of the main paper, indicating that the Event-distilled model does not rely on high-frequency texture as an inference cue. On the other hand, in the low- and mid-frequency bands ($0\pi$--$0.5\pi$), the gap between models narrows, indicating that the Event-distilled model's advantage is concentrated in the high-frequency range; in the low--mid range, both models are similarly affected because the Event-distilled model itself relies on these bands for edge extraction, as detailed in the noise frequency decomposition of Appendix~\ref{sec:ap_high_freq_noise_decomp}.

\subsection{Comprehensive Evaluation on ImageNet-C}
\label{sec:ap_high_freq_inc}

\begin{figure*}
\vspace{-1.5em}
\centering
\includegraphics[width=\linewidth]{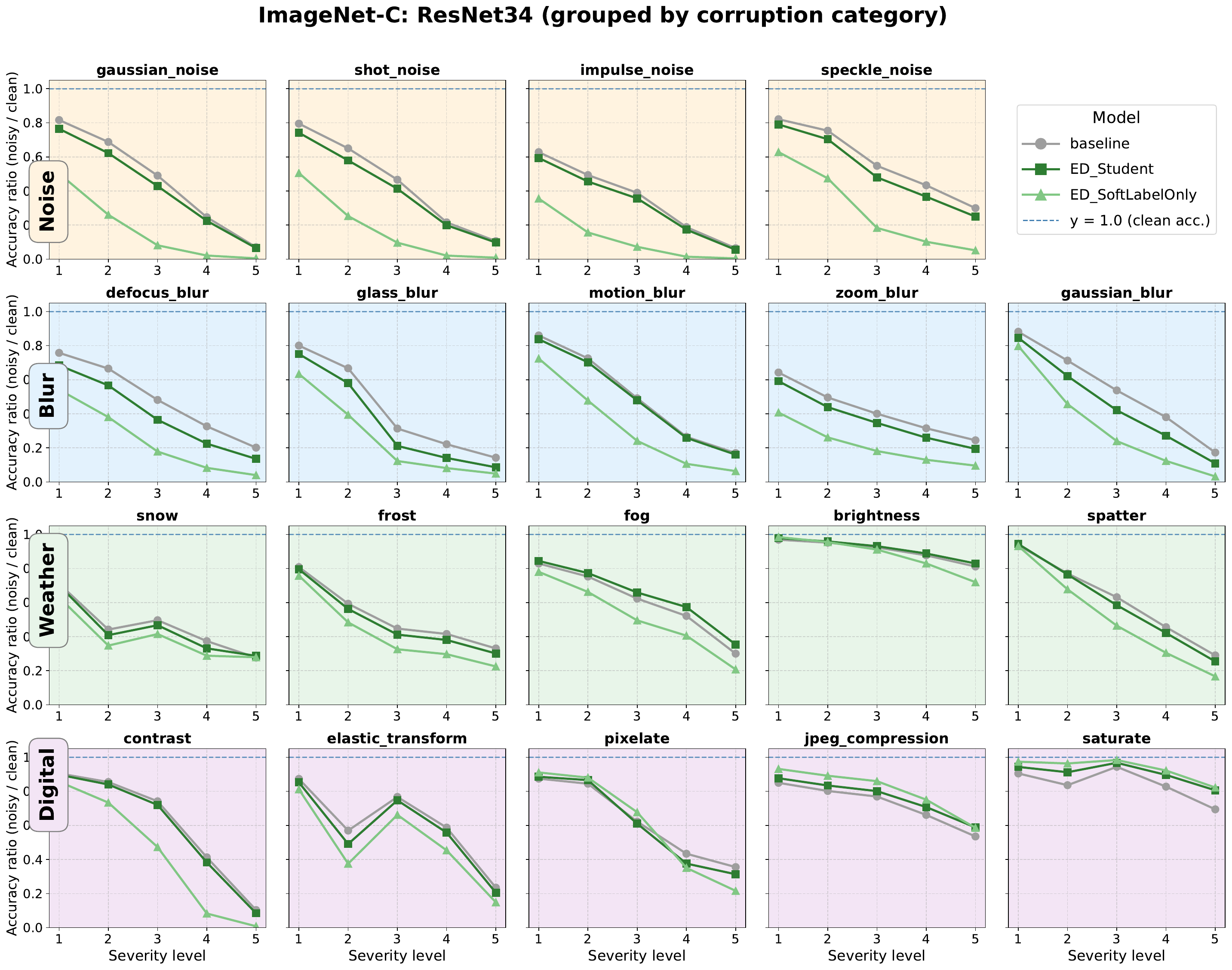}
\vspace{-1.1em}
\caption{Top-1 accuracy of ResNet-34 models (baseline, ED\_Student, ED\_SoftLabelOnly) against all 19 ImageNet-C corruptions, grouped into Blur / Digital / Noise / Weather. Horizontal axis is Severity ($1$--$5$).}
\label{fig:ap_fig_imagenet_c}
\end{figure*}

To comprehensively verify the spectral trade-off presented in Section~\ref{sec:5-2} of the main paper, Figure~\ref{fig:ap_fig_imagenet_c} reports the accuracy evaluation against all 19 corruptions comprising ImageNet-C. In the figure, corruptions are visualized in four groups: Blur, Digital, Noise, and Weather.

We summarize the observations that reinforce the main paper's claims for each group. The clearest gain appears on perturbations that selectively remove high-frequency texture on the image surface while largely preserving the macroscopic contours of objects. On Digital corruption jpeg\_compression and Weather corruption fog, the Event-distilled model consistently exceeds baseline, confirming that such distribution shifts are favorable to the Event-distilled model, which does not rely on texture. This supports the spectral-trade-off hypothesis of Section~\ref{sec:5-2} in the main paper.

On the other hand, on Blur (gaussian\_blur, defocus\_blur, motion\_blur, glass\_blur, zoom\_blur), the Event-distilled model's advantage is limited, and at high severities it can fall below baseline. Blur smooths not only high-frequency texture but the edges themselves, so for the Event-distilled model, which strongly depends on edges, blur attenuates its very inference cue. This observation indicates that the benefit of the Event-distilled model is restricted not to texture removal in general, but to texture removal under preserved edges, reflecting its strong dependence on the edge-based shape feature.

In contrast, on Noise (gaussian\_noise, shot\_noise, impulse\_noise, speckle\_noise), the Event-distilled model degrades to roughly the same accuracy as baseline or worse. Furthermore, on Digital corruptions that physically damage the structure itself (pixelate, contrast, elastic\_transform), at low severities the Event-distilled model maintains higher accuracy than baseline, but in regions where severity increases and the original edge continuity is lost, a reversal occurs and it falls below baseline. These vulnerabilities support the claim of Section~\ref{sec:5-2} of the main paper, that the Event-distilled model is strongly affected when the geometric structure and edge continuity on which it relies for inference are destroyed.

\subsection{Why Event-distilled Models are Fragile to Additive Noise}
\label{sec:ap_high_freq_noise_decomp}

\begin{figure*}
\vspace{-1.5em}
\centering
\includegraphics[width=\linewidth]{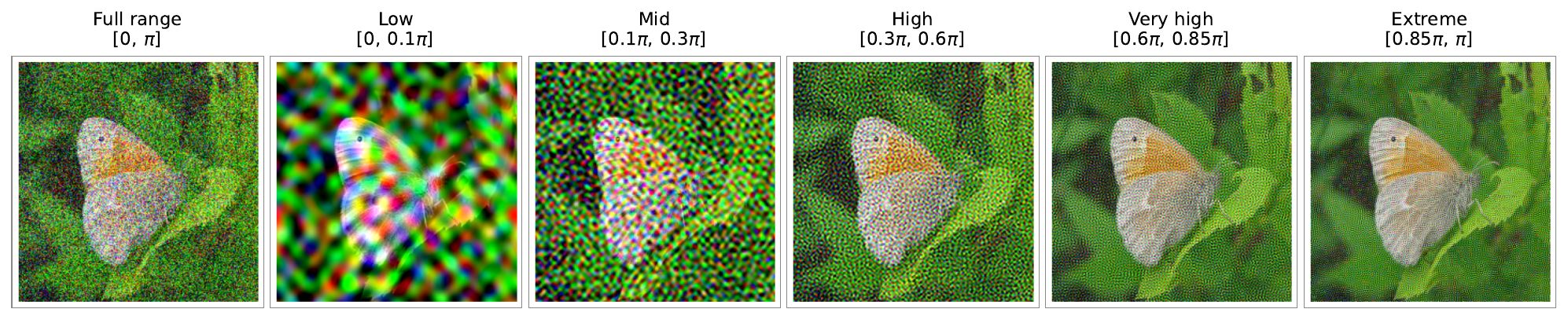}
\vspace{-1.1em}
\caption{An ImageNet validation image with additive Gaussian Noise (Full range, $[0, \pi]$), and the results of first decomposing the noise with band-pass filters into Low ($[0, 0.1\pi]$), Mid ($[0.1\pi, 0.3\pi]$), High ($[0.3\pi, 0.6\pi]$), Very high ($[0.6\pi, 0.85\pi]$), and Extreme ($[0.85\pi, \pi]$) and then adding each band's component to the image. The Full-range noise visually appears dominated by high-frequency granular components, but it carries energy across all frequency bands.}
\label{fig:ap_fig_noise_decomposition}
\vspace{-1em}
\end{figure*}

\begin{figure*}
\vspace{-1.0em}
\centering
\includegraphics[width=\linewidth]{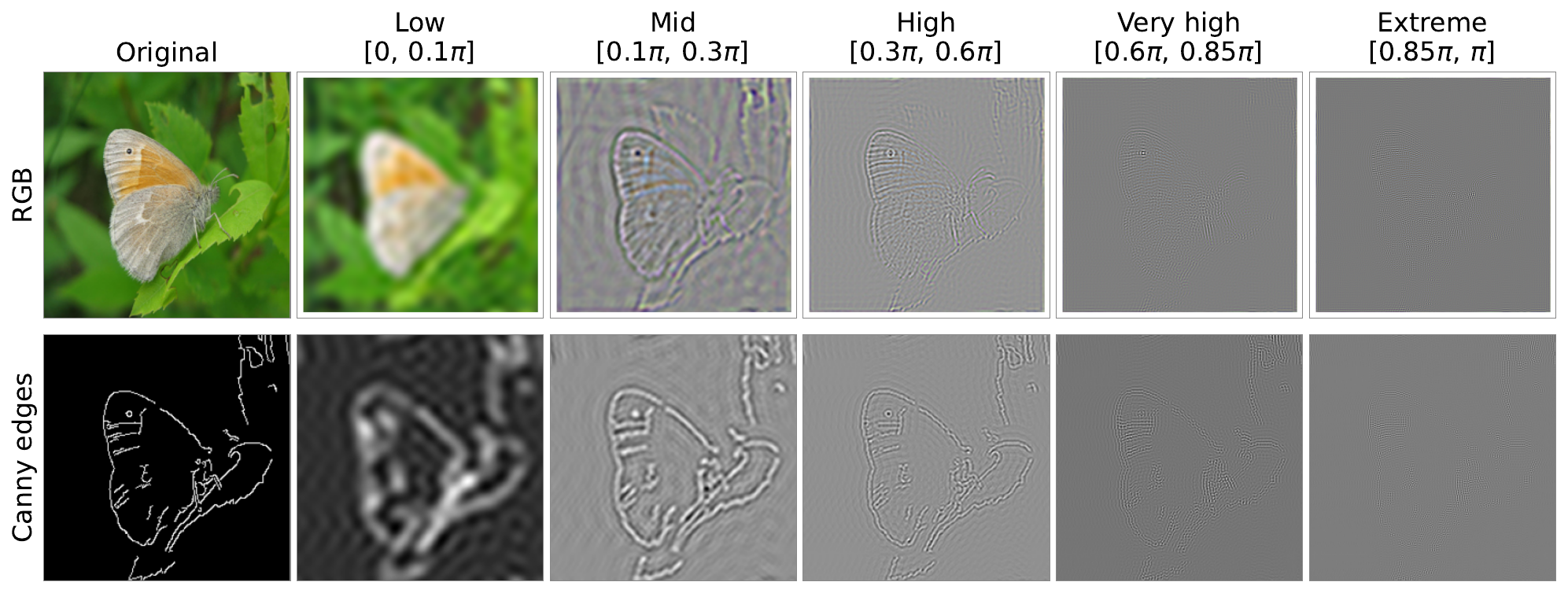}
\vspace{-1.1em}
\caption{The same ImageNet image (RGB, top row) and its Canny edge map (Canny edges, bottom row) decomposed by band-pass filters into the same bands as in Figure~\ref{fig:ap_fig_noise_decomposition}. The Original column shows the input that each row decomposes (top: the RGB image; bottom: the Canny-processed image). The energy of the RGB image (top) is concentrated in the Low--Mid bands ($[0, 0.3\pi]$). Even the Canny map (bottom), which consists solely of edge components, similarly concentrates its energy in the Low--Mid bands and carries virtually no energy in the Very-high or higher bands. In other words, edge information is not localized in the very-high band: the low-frequency band still carries broad edge components accompanied by gradual gradients, and the energy is biased toward the mid-frequency band.}
\label{fig:ap_fig_image_bandpass}
\vspace{-1em}
\end{figure*}

We further investigate the vulnerability to Noise corruptions confirmed in the previous subsection from the perspective of the noise's frequency characteristics. As pointed out by~\citet{yin2020noisetraining}, additive noises such as Gaussian Noise, Shot Noise, and Impulse Noise are white noise (or close to it) in the spatial domain, and behave as broadband noise that has energy across all frequencies in the Fourier domain, from low to high.

To visually confirm this property, we first decompose Gaussian noise with band-pass filters into different frequency bands and then add each band's component to the image; the resulting images are shown in Figure~\ref{fig:ap_fig_noise_decomposition}. As the figure shows, the Full-range noise injects strong noise components into the Low ($[0, 0.1\pi]$) and Mid ($[0.1\pi, 0.3\pi]$) frequency bands on which the Event-distilled model relies for inference. As shown in the top row of Figure~\ref{fig:ap_fig_image_bandpass}, the energy of the RGB image itself is also concentrated in the Low--Mid bands; not only do the steep edges constituting object contours appear predominantly in the mid-frequency band, but broad edge components accompanied by gradual gradients are also distributed in the low-frequency band. Consequently, the bands the Event-distilled model uses to extract correct edges and the bands the noise contaminates overlap heavily.

To further isolate the frequency distribution of edge information itself, we apply the same band-pass decomposition to a Canny edge map extracted from the RGB image; the result is shown in the bottom row of Figure~\ref{fig:ap_fig_image_bandpass}. Even though the input is composed solely of edges, its energy is similarly concentrated in the Low--Mid bands and is virtually absent from the Very-high and higher bands. This visually confirms that edge information does not have a one-to-one correspondence with very-high-frequency content: broad edge components are still distributed in the low-frequency band, while the energy is biased toward the mid-frequency band. This is consistent with our claim in this section that the Event-distilled model relies primarily on the mid-frequency band for edge information, and that contamination of the surrounding (Low--Mid) bands can therefore substantially impair performance.

The Event-distilled model has acquired a tendency to selectively exploit edge information while ignoring high-frequency texture (Section~\ref{sec:5-1} of the main paper); accordingly, it is strongly affected by contamination of the frequency bands that can disrupt either the local or the global aspects of edge information. This is evidence, from a different angle, that the Event-distilled model is not fragile to noise in general but rather that it strongly depends on the frequency components that carry edge information, and is consistent with the observation in Appendix~\ref{sec:ap_high_freq_fgsm} that performance differences disappear when noise intrudes into the frequency band on which the model relies.

\section{Spectral Statistics of Event vs. Natural Images}
\label{sec:ap_psd}

In this appendix, we corroborate the spectral trade-off hypothesis discussed in Section~\ref{sec:mechanism} of the main paper at the most upstream level --- the frequency statistics of the data itself --- by comparing the radial Power Spectral Density (PSD) of ImageNet (natural images) and N-ImageNet (event representation).

\begin{figure}[t]
\centering
\includegraphics[width=0.7\linewidth]{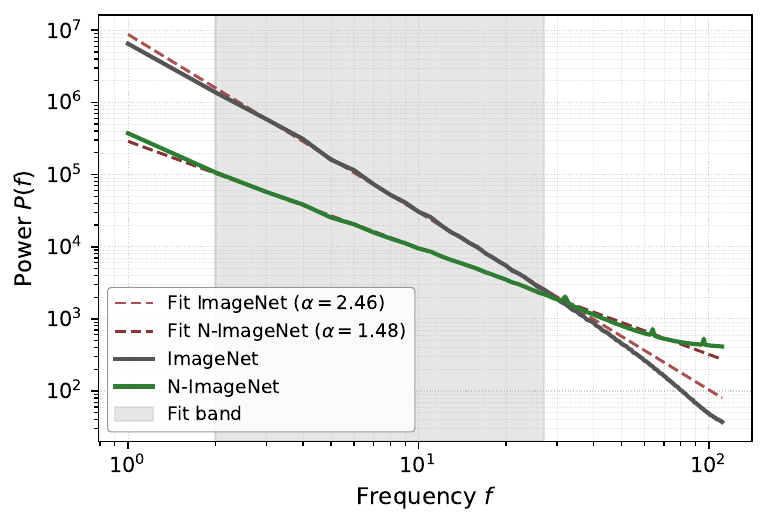}
\caption{Radial PSD of ImageNet (natural images) and N-ImageNet (event representation) plotted in log-log space. Linear fits are computed over the intermediate frequency band ($0.02 F \le f \le 0.25 F$, gray shading; $F$ denotes the maximum frequency) to estimate the power-law exponent $\alpha$ in $P(f) \propto 1/f^{\alpha}$. ImageNet yields $\alpha = 2.46$ ($R^2 = 0.9993$), broadly consistent with the empirically established $1/f^2$ statistics of natural images~\cite{van1996modelling}, while N-ImageNet yields $\alpha = 1.48$ ($R^2 = 0.9996$) --- a markedly shallower slope, indicating less concentration of energy at low frequencies.}
\label{fig:ap_psd}
\end{figure}

\paragraph{Setup.}
It is widely known that the PSD of natural images follows a power law $P(f) \propto 1/f^{\alpha}$ with $\alpha \approx 2$~\cite{van1996modelling}. We follow this convention and treat $\alpha$ as a positive exponent throughout this appendix. Following the standard procedure of \citet{van1996modelling, langer2000large}, we estimate $\alpha$ by linear regression in the log-log domain. We discard low-frequency components ($f < 0.02 F$) that are dominated by global structure and high-frequency components ($f > 0.25 F$) that are prone to noise and discretization effects, and fit only over the intermediate band, where $F$ denotes the maximum radial frequency.

\paragraph{Results and interpretation.}
As shown in Figure~\ref{fig:ap_psd}, ImageNet falls within the typical range of natural-image statistics ($\alpha = 2.46$), whereas N-ImageNet shows a slope shallower by roughly one point ($\alpha = 1.48$). This indicates that, in the N-ImageNet representation, the dominance of low-frequency components is weakened and the relative proportion of mid- to high-frequency components is correspondingly larger (i.e., the PSD is flatter).

This difference is consistent with the hardware design of the event sensor: an event sensor records only local changes in brightness, so smooth, spatially extended (low-frequency) textures and uniform backgrounds are not represented in its output. As a consequence, the frequency components that correspond to edges in the natural image (predominantly the mid- to high-frequency bands) are relatively emphasized, and the PSD slope becomes shallower. Note, however, that this does not mean edge information is localized exclusively in the high-frequency band: broad, diffuse edges accompanied by gradual gradients still carry energy in the low-frequency band as well (see Figure~\ref{fig:ap_fig_image_bandpass} in Appendix~\ref{sec:ap_high_freq_noise_decomp}). From the perspective of the spectral trade-off hypothesis discussed in Section~\ref{sec:mechanism} of the main paper, the N-ImageNet teacher learns from a distribution that is already richer in edge content; therefore, the student model that is distilled from it is steered, already at the input-statistics level, toward representations that do not depend on unstructured high-frequency texture and that selectively extract edge-based shape features (Section~\ref{sec:5-1} of the main paper). The first-layer filters in Figure~\ref{fig:filter_jpeg} and the spectral trade-off on ImageNet-C in Section~\ref{sec:5-2} can be understood as the consistent downstream manifestations of this upstream input-statistics difference at the representation and behavioral levels.

\section{Extended Discussion of Limitations}
\label{sec:ap_limitations}

This appendix expands on the limitations summarized in Section~\ref{sec:limitations} of the main paper.

While this work provides insights regarding the inductive bias derived from event data, several limitations exist, and they suggest future research directions. First, there is a limitation arising from the data-generation process of N-ImageNet used to train the teacher: the events are obtained by recapturing RGB images displayed on a monitor. This procedure successfully teaches the model the core characteristics of the event sensor (asynchronous response to spatial brightness gradients, i.e., edges, and high temporal resolution). On the other hand, due to the constraints of the monitor, the sensor's intrinsic high dynamic range (HDR) is not fully exploited; and because the captured content is planar imagery, complex motion parallax due to depth in the real world is missing. Second, since this validation focused on overcoming the texture bias intrinsic to CNNs, how Event distillation transforms the internal representation of Vision Transformers (ViT)~\cite{dosovitskiy2021vit}, which are reportedly more shape-biased to begin with, is not yet known. Finally, although the validation in this work is limited to image classification, the macroscopic edge representation acquired here may function as a useful prior for dense-prediction tasks such as semantic segmentation; verifying these aspects with more native event datasets is an important step for future work.


\end{document}